\documentclass[journal]{IEEEtran}
\usepackage[utf8]{inputenc}
\usepackage{amsmath}
\usepackage{multirow}
\usepackage{epstopdf}
\usepackage{enumitem}
\usepackage{color}
\usepackage{graphicx}
\usepackage{booktabs}
\usepackage{siunitx}
\usepackage{cite}

\usepackage{array,multirow,multicol,rotating,tabularx,ragged2e,booktabs}

\ifCLASSINFOpdf
\else
\fi
\usepackage[dvipsnames]{xcolor}
\usepackage{booktabs}
\usepackage{amsmath}
\usepackage{array,multirow,multicol,rotating,tabularx,ragged2e,booktabs}

\newcommand{\NW}[1]{\textcolor{black}{#1}}
 
 \newcommand{\SB}[1]{\textcolor{black}{#1}}
 
\newcommand{\bc}[1]{\textcolor{blue}{#1}}

\begin{document}

\title{Learned 3D Shape Representations Using Fused Geometrically Augmented Images: Application to Facial Expression and Action Unit Detection}

%
%
%
 
 \author{Bilal~Taha, Munawar~Hayat,Stefano Berretti, ~and~Naoufel~Werghi}

\markboth{}
{}
%



\maketitle


\begin{abstract}
\NW{This paper proposes an approach to learn generic multi-modal mesh surface representations using a novel scheme for fusing texture and geometric data. Our approach defines an inverse mapping between different geometric descriptors computed on the mesh surface or its down-sampled version, and the corresponding 2D texture image of the mesh, allowing the construction of fused geometrically augmented images (FGAI). This new fused modality enables us to learn feature representations from 3D data in a highly efficient manner by simply employing standard convolutional neural networks in a transfer-learning mode. In contrast to existing methods, the proposed approach is both computationally and memory efficient, preserves intrinsic geometric information and learns highly discriminative feature representation by effectively fusing shape and texture information at data level. The efficacy of our approach is demonstrated for the tasks of facial action unit detection and expression classification. The extensive experiments conducted on the Bosphorus and BU-4DFE datasets, show that our method produces a significant boost in the performance when compared to state-of-the-art solutions.}
\end{abstract}

\begin{IEEEkeywords}
\NW{Mesh surface, fused geometrically augmented images, convolution neural networks, expression recognition.}
\end{IEEEkeywords}

\IEEEpeerreviewmaketitle

\section{Introduction}
\IEEEPARstart{C}{ompared} to 2D photometric images, 3D data in the form of mesh surfaces provides more information and is invariant to illumination, out-of-plane rotations and color variations. Further, it provides geometric cues, which enable better separation of the object of interest from its background. Despite being more promising and information-rich, the focus of previous research on representing 3D data has been to carefully design hand-crafted methods of feature description. While automatically learned feature representations in terms of activations of a trained deep neural network have shown their superiority on a number of tasks using 2D RGB images, learning generic shape representations from 3D data is still in its infancy.

Among the different application contexts where shape representations have consolidated their relevance, face analysis is indubitably one of the topics of more active 3D research. On the one hand, this is motivated by the fact that 3D face data can complement 2D images to enhance face analysis in difficult conditions as in the case of facial expressions, occlusions, pose and illumination changes, or in specific tasks where 2D information alone is not sufficient as for face spoofing. On the other hand, 3D face research is boosted by the availability of an increasing number of devices that can be easily used to acquire the face in 3D either at high-resolution, using 3D static/dynamic scanners, or at low-resolution, with 3D cameras like Kinect.
A specific face analysis task, which is attracting increasing interest, is that of recognizing facial expressions from 3D static or dynamic data. 
In fact, facial expressions are one of the most important ways of person-to-person non-verbal communication, by which humans convey, either deliberately or in an unconscious way, their emotional state.
The way humans perform and perceive expressions has been studied for long time, with the seminal works by Ekman et al.~\cite{ekman:1992}, showing that human facial expressions can be categorized into six \textit{prototypical} classes, namely, \textit{angry}, \textit{disgust}. \textit{fear}, \textit{happy}, \textit{sad}, and \textit{surprise} that are invariant across different cultures and ethnic groups.
Later studies have also shown that it is possible to think of expressions as the facial deformations induced by the movement of one or more muscles; such atomic deformations have been classified according to a Facial Action Coding System (FACS)~\cite{ekman:1978}, where a code is used to identify an Action Unit (AU) corresponding to the effect of individual or groups of muscles.
\begin{figure*}
\centering
\includegraphics[width=0.9\linewidth]{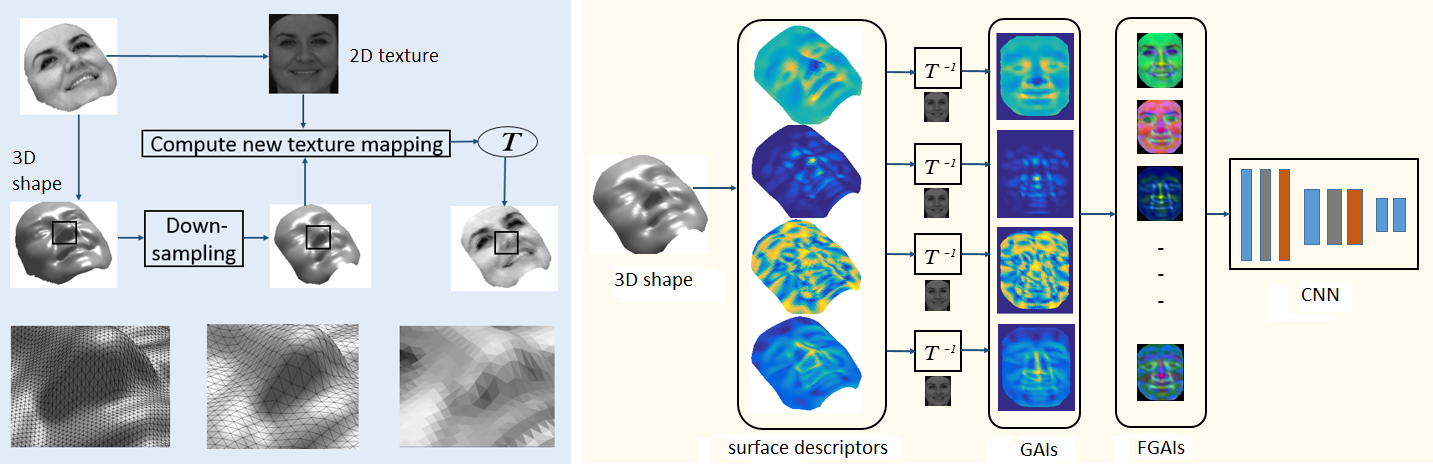} \\
\small{(a)   \hspace{0.5\linewidth}   (b)}
\caption{(a) Computation of a new texture mapping transformation $\mathcal{T}$ between the down-sampled mesh model of the face and its texture image, both derived from the original face model.
(b) A variety of geometric descriptors are computed on the down-sampled 3D face mesh surface. These are mapped to the texture 2D image using the inverse texture mapping function $\mathcal{T}^{-1}$. From the constructed images, dubbed \textit{geometry-augmented images} (GAIs), we derive combinations of image triplets arranged into three-channel images. These latter images, dubbed \textit{fused geometry-augmented images} (FGAIs) are used with different CNN models, thus learning highly discriminative feature representations.}
\label{fig:blocdiagram}
\end{figure*}

In this paper, we propose an original approach to extend the application of deep learning solutions to 3D data given in the form of triangular meshes. This is obtained by developing on the idea of establishing a mapping between the 3D mesh domain and the 2D image domain. Existing mapping solutions directly generate 2D photometric images by flattening the 3D model to the image plane, or representing it by different 2D views. But, in doing so, descriptors are computed on the 2D generated images, thus losing most of the real geometry of the 3D shape.
Different from previous works, we propose to directly capture the geometric information from 3D mesh surface in terms of a set of local geometric descriptors. The extracted geometric information is then fused in our proposed geometrically augmented 2D images which can efficiently be used in conjunction with state-of-the-art CNN models.
\NW{Moreover, our method makes it possible to compute the geometric descriptors on a down-sampled version\footnote{We will use the terms down-sampled and compressed interchangeably.} of the mesh-model allowing a considerable gain in efficiency without compromising the performance.}

Compared with existing methods, the proposed approach faithfully preserves the intrinsic geometric structure in terms of local descriptors, is computationally efficient, and does not require memory intensive tensor representations.
As shown in the block diagram of Fig.~\ref{fig:blocdiagram}, our proposed framework jointly exploits the 3D shape and 2D texture information. First, we perform a down-sampling on the facial surface derived from the full 3D face model. Subsequently, we compute the new texture mapping transformation $\mathcal{T}$ between the texture image and the compressed mesh. Afterward, we extract local shape descriptors in terms of curvatures, shape index and local depth on the 3D shape (details are given in Sect.~\ref{sect:Local-Geometric}). A novel scheme is then proposed to map the extracted geometric descriptors onto 2D textured images, using the inverse of the texture mapping transform $\mathcal{T}$ (see Sect.~\ref{sect:3D-to-2D}). The mapping preserves the shape information, while compactly encoding the geometric description in 2D. We dubbed these images \textit{geometry-augmented images} (GAIs).
It is relevant to remark here that, in the proposed mapping, we assume the existence of a 2D texture image, which is in correspondence with the triangulated mesh via standard texture-mapping. In this respect, our solution can also be regarded as a multi-modal 2D+3D solution, where the 2D texture data, at the same time, is required to enable the mapping, and also constitutes an additional feature that can be early fused with the 3D data in a straightforward way.
\NW{The GAIs are then combinatorially fused to generate multiple three-channel images, which are used to learn highly discriminative feature representations. We dubbed these images \textit{fused geometry-mapped images} (FGAIs).
The effectiveness of the proposed learned feature representation scheme is demonstrated through extensive experiments on the Bosphorus dataset, for the tasks of facial expression classification and AU detection, and on the BU-4DFE dataset, for the tasks of static and dynamic facial expression recognition.} In summary, the original contributions of this work are:\\
1-We propose a new scheme, which maps the geometric information from compressed 3D meshes onto 2D textured images. The proposed scheme provides us with a mechanism to simultaneously represent geometric attributes from the mesh-model alongside with the texture information. The proposed mapped geometric information can be employed in conjunction with a Convolutional Neural Network (CNN) for feature learning from multi-modal (2D and 3D) data.\\
2-We provide a highly discriminative representation of 3D data and texture data in terms of activations of a trained CNN model. Compared to existing learned feature representation schemes for 3D data, the proposed method is both memory and computation efficient as it does not resort to expensive tensor-based or multi-view inputs.\\
 3-The proposed scheme allows us to intrinsically fuse texture and shape information at data-level by mapping 3D geometric information in terms of local descriptors onto 2D textured images. Compared with other score or decision level fusion schemes, the proposed approach jointly learns to fuse 2D and 3D information at the data level. Such low-level data fusion has been shown to be more effective compared with the high-level fusion of scores or decisions~\cite{ross2003}.\\
4-We propose a novel geometric descriptor, called \textit{local depth}, which effectively encodes the depth value of a point on the 2D manifold, within a local neighborhood (details are given in Sect.~\ref{sect:Local-Geometric}).

To the best of our knowledge, we are the first to propose such 2D and 3D information fusion for textured-geometric data analysis.
The approach by Li et al.~\cite{lin2017} is the closest one to our proposed solution. However, our method presents two fundamental differences with respect to their work. First, Li et al.~\cite{lin2017} separately encoded texture and geometric information, and dedicate a sub-network to each descriptor. The related output features go into a subsequent feature-level fusion network. In contrast, in our approach, the texture and the geometric information are fused at the data level by mapping the geometric descriptors onto texture images, then rendering multiple three-channel images, which are fed as input to the CNN model. Second, the geometry maps in Li et al.~\cite{lin2017} are obtained by computing the geometric attributes on the face mesh model, then displaying and saving them as 2D images. In our method, we rather establish a correspondence between geometric descriptors computed on 3D triangular facets\footnote{The term facet will be used to refer to a triangular face of the mesh.} of the mesh and pixels in the texture image. Specifically, in our case, geometric attributes are computed on the mesh at each facet and then mapped to their corresponding pixels in the texture image using the newly proposed scheme. This yields to a sort of multi-spectral image, where each pixel is an aggregation of texture and geometric information in the form of local descriptors. Such aggregation allows us to encode facial shape and texture in a different multi-channel image representation, hence offering a new data augmentation mechanism. \NW{Third, our method is computationally more-efficient as we compute the geometric attributes on a compressed facial mesh surface.}

\NW{A recent work close to ours, which is worth to mention, is the conformal mapping method proposed by Kittler et al.~\cite{kittler2016}. Like our proposed solution, this method maps 3D information to 2D face images to adapt the data representation to CNN processing. However, there are three fundamental differences between their approach and ours. First, in~\cite{kittler2016}, the mapping is performed using a single generic 3D morphable face model (3DMM) to the 2D face image, whereas in our solution we map each actual face model of a subject to its corresponding 2D face image. Second, the geometric information mapped from the 3DMM to the 2D image is given by the point coordinates which are just an estimate of the actual values, obtained from the 3DMM by landmark-based model fitting. In our work, instead, we map the actual point coordinates, in addition to other varieties of shape descriptors, from each face model to its corresponding image. Third, regarding the scope, the work of Kittler et al.~\cite{kittler2016} deals with 2D face recognition, while our work addresses facial expression and AU recognition using 3D face images.
Finally, we also mention the works of Zhu et al.~\cite{zhu2016} and Sinha et al.~\cite{sinha:2016}. In~\cite{zhu2016}, similarly to~\cite{kittler2016}, a 3DMM is fit to a 2D image, but for face image frontalization purposes. The face alignment method employs a cascaded CNN to handle the self-occlusions. In~\cite{sinha:2016}, a spherical parameterization of the mesh manifold is performed allowing to have a two-dimensional structure that can be treated by conventional CNNs.}

The remainder of the paper is organized as follows. Existing methods for learning representations from 3D shape data and facial image analysis are discussed in Sect.~\ref{sect:Related-work}. Then, in Sect.~\ref{sect:3D-Learned-Feature}, we provide a description of our proposed scheme, followed by extensive experiments in Sect.~\ref{sect:Experimental-Results} to empirically validate the efficacy of the proposed method. Discussion and conclusions are given in Sect.~\ref{sect:conclusions}.

\section{Related work}\label{sect:Related-work}
In the following, we summarize the works in the literature that are more closed to our proposed method. In particular, first, we revise methods that have been used to represent the shape of 3D models surface (Sect.~\ref{sect:3d-shape-representation}); 
then, we report on the approaches that addressed facial expression recognition and AU detection from 3D static and dynamic data (Sect.~\ref{sect:3d-expression-recognition}).

\subsection{3D Shape Representation}\label{sect:3d-shape-representation}
Most of the research on 3D object classification uses carefully hand-crafted descriptors.
Guo et al.~\cite{guo2016comprehensive} presented a comprehensive survey and performance comparison of such descriptors, whereby the authors categorized these traditional descriptors into two groups:
\emph{(i)} descriptors based upon histograms of spatial distributions (\emph{e.g.}, point-clouds). \emph{(ii)} descriptors based upon histograms of local geometric characteristics (\emph{e.g.}, surface normals and curvatures). 

In contrast to learned feature representations, a significant limitation of the traditional hand-crafted descriptors is their lack of generalization across different domains and tasks. These hand-crafted descriptors have been recently outperformed by features learned from raw data using deep CNN.
Feature learning capabilities of CNNs from RGB images have been demonstrated in a number of challenging Computer Vision tasks such as object classification and detection, scene recognition, texture classification and image segmentation~\cite{krizhevsky2012imagenet,redmon2016you}. Features learned on a generic large scale dataset such as ImageNet have been shown to generalize well across other tasks, \emph{e.g.}, fine-grained classification and scene understanding~\cite{sharif2014cnn}. Due to their excellent generalization capabilities and impressive performance gain, learned feature representations are believed to be superior compared with traditional hand-crafted features. Despite their success on 2D images, research on learning features from 3D geometric data is still in its infancy compared with its 2D counterpart. Below, we provide an overview of existing 3D deep learning methods.

In one of the earliest works, CNNs and recursive neural networks (RNNs) have been jointly trained on RGB-D data~\cite{socher2012convolutional}.
Gupta et al.~\cite{gupta2014learning} learned geocentric embedding, which encodes height above ground and angle with gravity for depth images. Eitel et al.~\cite{eitel2015multimodal} first separated the color (RGB) and depth (D) information through a CNN followed by late-fusion for RGB-D object detection. Compared with RGB-D data, 3D data in the form of mesh model provides complete and more structured shape information. New methods have been developed to represent and learn features from such data. These methods are discussed below.

\NW{Approaches for learning features from 3D data can be divided into two categories: \textit{volumetric} approaches and \textit{manifold} approaches. The first category treats 3D volumetric data and encompasses basically two paradigms: volumetric CNNs and multiview CNNs. Volumetric CNNs process 3D data in its raw format (\emph{e.g.}, a volumetric tensor of binary/real-valued voxels~\cite{wang2015}. Unlike 2D images, where each pixel carries meaningful information, only the voxels corresponding to the object surface and boundaries are helpful. Volumetric representation based CNNs are therefore memory intensive and inefficient. Recent works addressed this problem by proposing architectures operating on a cloud of points, while respecting the permutation invariance of points in the input~\cite{qi2017}.
Multiview CNNs paradigm~\cite{crqi2016} extends 2D CNNs to 3D data by synthetically rendering multiple 2D images across different viewpoints of a given 3D point-cloud. These multiple images are then fed as inputs to CNNs, followed by a fusion scheme to get a single entity representation of the 3D shape. These multi-view representations have shown superior performance compared with volumetric approaches. However, a limitation of multi-view scheme is that 3D geometric information is not fully preserved in rendering images from 3D data.}

\NW{Manifold approaches operate on mesh surfaces, which serve as a natural parametrization to 3D shapes; but learning using CNNs is a challenging task in such modality. Current paradigms to tackle this challenge either adapt the convolutional filters to mesh surfaces or learn spectral descriptors defined by the Laplace-Beltrami operator. Boscaini et al.~\cite{boscaini:2015} proposed a generalization of CNNs to non-Euclidean domains for the analysis of deformable shapes based on localized frequency analysis. Masci et al.~\cite{masci:2015} extended the CNN paradigm to non-Euclidean manifolds by using a local geodesic system of polar coordinates to extract ``patches'' on which  geodesic convolution  can be computed. Seong et al.~\cite{seong:2017} introduced a geometric CNN (gCNN) that deals with data representation over a mesh surface and renders pattern recognition in a multi-shell mesh structure. Wang et al.~\cite{wang:2017b} built a hash table to quickly construct the local neighborhood volume of eight sibling octants that allow an efficient computation of the 3D convolutions of these octants in parallel.}

\subsection{Facial Expression Recognition}\label{sect:3d-expression-recognition}
Most of the work on facial expression recognition and AU detection has been done using 2D data. A survey of these works appeared in~\cite{Caleanu2013}. Here, we review the methods developed for 3D data only. We can broadly categorize 3D facial analysis methods into two groups: \textit{feature-based} and \textit{model-based}.
Feature-based methods extract geometric descriptors either holistically or locally from the 3D facial scans. For example, Zhao et al.~\cite{Zhao2010}  detect facial landmarks on a given 3D face. Local geometric and texture features were then extracted around the detected landmarks, and used to represent the 3D facial scan. Similarly, Maalej et al.~\cite{Maalej2011} represented a facial scan in terms of local surface patches extracted around $70$ facial landmarks. Geodesic distance on the Riemannian geometry is utilized as a metric to compare the extracted patches. A number of other works represented 3D scans using either local or holistic geometric descriptors. Examples include distances between 3D facial landmarks~\cite{Tang2008}, distances between locally extracted surface patches~\cite{LI2015}

Model-based approaches first establish a dense point-to-point correspondence between a query mesh and a generic expression deformable mesh using rigid and non-rigid transformation techniques. The transformation parameters are then used as representations of the query mesh for classification. Non-rigid facial deformations are characterized by a bilinear deformable model in~\cite{Mpiperis2008}. The shape of a scan with facial expression is decomposed into neutral and expression parts in~\cite{Gong2009}. The expression part of the decomposed scan is then employed for encoding the facial scan. Some works combine the strengths of both feature-based and model-based techniques. For example, Zhen et al.~\cite{Zhen2016} first segment the face into multiple regions based upon muscular movements. Geometric descriptors are then extracted from these regions, followed by a fusion scheme weights to optimally combine decisions from different regions.

\NW{More recently, following the release of the BU-4DFE database~\cite{yin2008}, several works were interested in the dynamic face recognition.
Sun and Yin~\cite{yin2008} proposed a deformable heuristic model representing both static and dynamic information with spatiotemporal descriptors, followed by a 2D hidden Markov Model (HMM) for the video expression classification. Fang et al.~\cite{Fang2011-ICCVW}, used the Mesh-HOG for matching faces and dynamic Local Binary Patterns (LBP) descriptor as an additional feature to capture temporal indices; an SVM was used for the classification. In~\cite{SANDBACH2012-IVC}, Sandbach et al. extracted 3D motion primitives, namely, Free-Form Deformation (FFD), between neighboring 3D Frames, then a GentleBoost classifier and HMMs have been adopted for recognizing the expression. Reale et al.~\cite{real2013}  proposed a dynamic regional joint histogram of curvature and polar angles; then LBP-features are extracted and used in nearest-neighbor classifier. Berretti et al.~\cite{Berretti2013}, modeled facial deformation with mutual distances between facial landmarks, and used two HMMs for the expression classification. Ben Amor et al.~\cite{Amor2014} presented collections of radial curves to describe the face, from which they derived quantified motion cues across successive 3D frames, using Riemannian geometry tools. LDA and HMM were subsequently used to classify expressions. As in~\cite{Berretti2013}, Xu et al.~\cite{Xue2015} detected facial landmarks to construct localized temporal depth maps from which 3D-DCT were derived to form  dynamic face signatures fed later into a  nearest-neighbor classifier. A similar region-wise description paradigm was proposed by Zhen et al.~\cite{Zhen2016}, where a variety of geometric descriptors were derived to weighted muscular areas, and used with an SVM and a muscular movement model (MMM) for the recognition. These same classifiers have been used in a subsequent work in~\cite{Zhen2018}, but adopting a spatial deformation encoded with Dense Scalar Fields, then using temporal filtering to amplify facial dynamic. Yao et al.~\cite{Yao2018} proposed a textual and geometric diffusion facial representation using a scattering operator. A multiple learning of the kernel (MKL) was used to combine the contributions of different channels.
Another category of methods adopted a random forest classifier~\cite{meguid2018,Dapogny2018}. Meguid et al.~\cite{meguid2018} used PCA features, derived from the dataset, and collection of binary random forests coupled with an SVM classifier.
Dapogny et al.~\cite{Dapogny2018} used a mixture of geometric and HOG features with a Greedy Neural Forest Classifier.}
\section{3D Learned Feature Representation}\label{sect:3D-Learned-Feature}
\SB{Our proposed 3D learned feature representation is constituted by several steps as illustrated in Fig.~\ref{fig:blocdiagram}. In the following, we deepen each stage as follows: first, four local geometric descriptors are computed on the triangular mesh manifold as explained in Sect.~\ref{sect:Local-Geometric}; then, such descriptors are projected to the 2D domain using an original solution that combines mesh resampling and inverse texture mapping (see Sect.~\ref{sect:3D-to-2D}). Triplets of GAIs originated from the 3D-to-2D descriptors mapping, and also from the gray-level texture image associated to the mesh, are then fused into the FGAIs and used as input to a DCNN architecture (Sect.~\ref{sect:Mapped-Image}); the output activations of the DCNN model generate a highly discriminative feature representation, which is used in the subsequent classification stage (Sect.~\ref{sect:Classification})}.

\subsection{Local Geometric Descriptors}\label{sect:Local-Geometric}
Given a triangular mesh manifold obtained from a 3D point-cloud, we compute four local geometric descriptors, namely, \textit{mean curvature} (\textit{H}), \textit{Gaussian curvature} (\textit{K}), \textit{shape index} (\textit{SI}), and the \textit{local-depth} (\textit{LD}).
These local descriptors can be computed efficiently and complement each other by encoding different shape information. In addition to these, we also consider the \textit{gray-level} (\textit{GL}) of the original 2D texture image associated to the mesh. 
For completeness, we provide below a brief explanation of the geometric descriptors.

The Gaussian curvature ($K$), the mean curvature ($H$), and the shape index ($SI$) are computed using the following equations~\cite{topon2006}:
\begin{eqnarray}
K = \lambda_1 \lambda_2 , \;
H = \frac{( \lambda_1 + \lambda_2)}{2} , \; 
SI = 1/2 - \frac{1}{\pi}  ( \frac{\lambda_1 + \lambda_2}{\lambda_2 + \lambda_1} ) \; .
\end{eqnarray}

\noindent
Here $\lambda_1$ and $\lambda_2$ are the principal curvatures determined by the roots of the following quadratic equation, for the unknown $k$~\cite{topon2006}:
\begin{equation}
\begin{vmatrix}
L-kE &  M-kF \\
M-kF & N-kG
\end{vmatrix}
= 0 \; ,
\end{equation}

\noindent
where $(E,F,G)$ and $(L,M,N)$ are, respectively, the coefficients of the first and the second fundamental form, computed at the given point $(x,y,z)$ on the surface. These coefficients are defined by:
\begin{eqnarray}
E = 1 + (\frac{\partial z}{\partial x})^2, \;\;\;\;\;\;\;\;\;\;\;\;\;\;\;\;\;\;\;\;\;\;\;\;\;\;\; \;\;\;\;\;
F = \frac{\partial z}{\partial x} \frac{\partial z}{\partial y} \; \\
\; \;\;\; \; \;\;\;G = 1 + (\frac{\partial z}{\partial y})^2,\;\;\;\;\;\; \;\;\; \;\;\;
M = \frac{\partial z}{\partial x}\frac{\partial z}{\partial y}/(EG-F)^2 \; \\
N = \frac{\partial^2 z}{\partial y^2}/(EG-F)^2, \;\;\;\;\;\;\;
L  =  \frac{\partial^2 z}{\partial x^2}/(EG-F)^2 
\end{eqnarray}

\begin{figure}[t]
\centering
\includegraphics[width=0.4\linewidth]{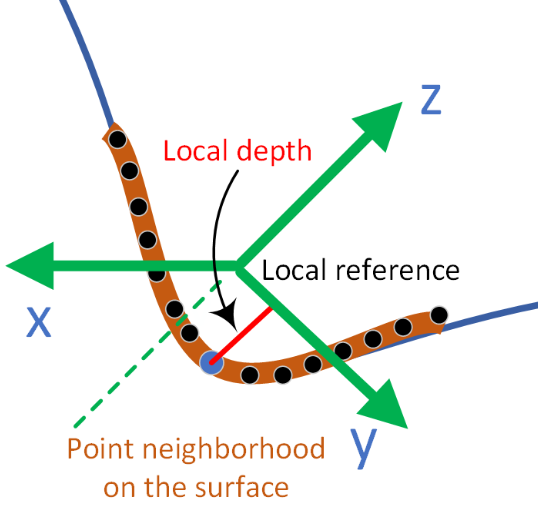}
\caption{The local depth is the algebraic distance (red segment in the figure) from a point on the surface to the main plane $(x,y)$ of the local reference spanning a local neighborhood  (the sectional brown curve).}
\label{fig:localdepth}
\end{figure}

The \textit{local-depth} (LD) is our newly proposed descriptor, which represents the depth value of a point in a local reference system attached to its neighboring points on the manifold. \NW{We define the neighborhood as the set of points encompassed by the geodesic disc of radius $3e$, where $e$ is the average edge length of the triangles of the mesh surface.}
The following steps are performed to compute the local depth at a given point on the mesh manifold. First, the local canonical reference system is computed at the point neighborhood (Fig.~\ref{fig:localdepth}). This reference is defined by the center of mass of the points within that neighborhood and the three eigenvectors of the Principal Component Analysis (PCA) of their covariance matrix. The local depth is the $z$ coordinate in the local reference obtained by computing the algebraic distance between the point and the plane $(x,y)$ \SB{spanning the first two principal components}, \NW{and having as normal the least principal component.}
\subsection{3D to 2D Mapping}\label{sect:3D-to-2D}
The core procedure of the proposed method is constituted by the mapping that we establish between the 3D and the 2D domain. This process assumes to have a 3D triangular mesh with its 2D texture image as inputs, and produces as output a set of 2D images representing the different surface descriptors computed on the mesh surface (\emph{i.e.}, GAIs).

Given a triangular mesh originated from a face scan, its vertices can be regarded as a cloud of scattered points. These points are initially projected onto the plane spanned by the main orientation of the face. This yields the depth function $z=f(x,y)$ defined on the scattered set of points. The function is then interpolated and re-computed over a regular grid constructed by a uniform down-sampling of order $k$, \NW{where $k$ defines the sub-sampling step.}
The 2D Delaunay triangulation computed over the achieved regular points produces a set of uniform triangular facets. We complete them with interpolated $z$ coordinate to obtain a 3D down-sampled regular mesh. This is illustrated in the middle face portion at the bottom of Fig.~\ref{fig:blocdiagram}~(a). In this procedure, we also compute, for each vertex in the new re-sampled mesh, its nearest neighbor in the original mesh.
\NW{A sub-sampling of step $k$ produces approximately a compression ratio (the original data over the compressed data)  of $k$ for both the facets and the vertices.}

According to the initial hypothesis, the original 3D mesh has an associated texture image. The mapping between the mesh and the texture image is established by the usual texture mapping approach, where the vertices of each triangle (\emph{i.e.}, also called facet in the following) on the mesh are mapped to three points (\emph{i.e.}, pixels) in the image. It is evident that the projection and re-sampling step, as illustrated above, break such correspondence that so needs to be re-established. Therefore, in the next stage, we reconstruct a new texture mapping between the 2D face image and the newly re-sampled mesh vertices.  For each vertex in the re-sampled mesh, we use its nearest neighbor in the original mesh, computed in the previous stage, to obtain the corresponding pixel in the original image via the texture mapping information in the original face scan. This new obtained texture mapping transformation ($\mathcal{T}$ in Fig.~\ref{fig:blocdiagram}) between the original image and the new re-sampled facial mesh allows us to map descriptors computed on the facial surface, at any given mesh resolution, to a 2D \textit{geometry-augmented image} (GAI), which encodes the surface descriptor as pixel values in a 2D image structure.

\begin{figure}[t]
\centering
\includegraphics[width=\linewidth, height=3cm]{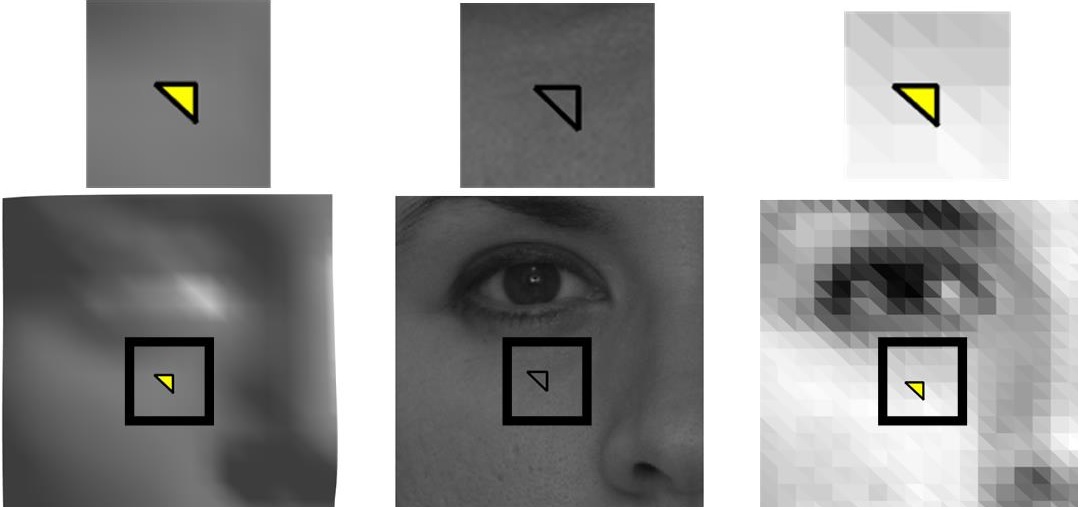}
\footnotesize{ (a) \hspace{2.3cm} (b) \hspace{2.3cm}  (c)}
\caption{(a) A triangular facet in a down-sampled face mesh; (b) The texture mapping associates the vertices of the facet in (a) to three pixels in the texture image forming thus a triangle; (c) All the pixels confined in the  triangle in (b) are assignrf their mean gray level (this results in a quite evident effect of triangular tessellation on the texture image). Ultimately, the texture in (c) constitutes the image counterpart of the mesh down-sampling.}
\label{fig:subsamplegray}
\end{figure}

\begin{figure}[b]
\centering
\includegraphics[width=\linewidth, height=3cm]{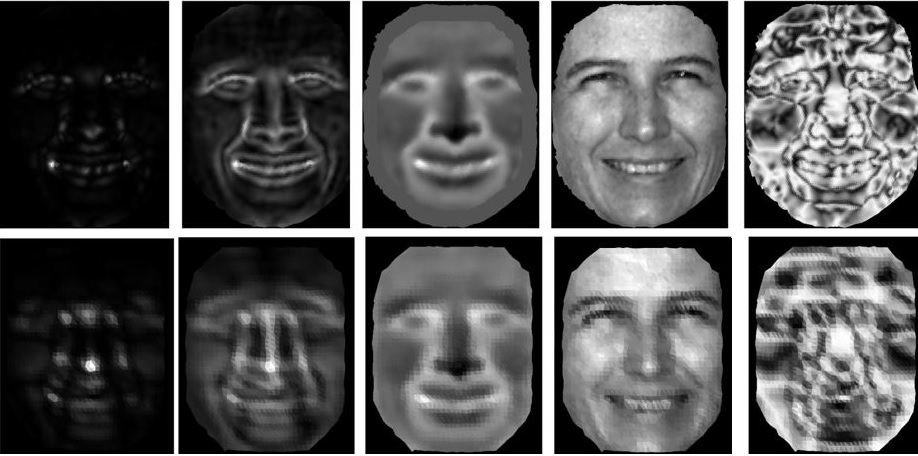} \\
\footnotesize{ K \hspace{1.2cm} {H} \hspace{1.1cm} {LD} \hspace{1.1cm} {GL} \hspace{1.2cm} {SI}}
\caption{GAIs obtained by 2D mapping of the descriptors computed on the original mesh (top row), and its down-sampled version (bottom row). From left-to-right: Gaussian curvature (K), mean curvature (H), local-depth (LD), gray level (GL), and shape index (SI). \NW{The compressed mesh is obtained with a sub-sampling step of 3.5. For this example, the number of facets/vertices in the original and compressed mesh are (9558/4873) and (2799/1459), respectively.}}
\label{fig:hrlr}
\end{figure}

\NW{We note here that, in order to keep consistent correspondence between the texture information and the geometric descriptors, we down-sample the original texture images to bring it at the same resolution of its GAI counterpart. We do this as follows: we take the vertices of each facet in the down-sampled facial mesh (Fig.~\ref{fig:subsamplegray}~(a)), and we compute, using $\mathcal{T}^{-1}$ their corresponding pixels in the original image (Fig.~\ref{fig:subsamplegray}~(b)). These three pixels form a triangle in the original image; we assign to all pixels that are confined in that triangle the average of their gray-level values (Fig.~\ref{fig:subsamplegray}~(c)).}
Figure~\ref{fig:hrlr} depicts the GAIs obtained by mapping the face descriptors computed on the original mesh (top row), and its down-sampled counterpart, at a compressed ratio of \NW{3.5}.
As the facial expressions affect the face shape at a macro-level, we advocate that the down-sampling, while significantly reduces the computational complexity, should not severely compromise the performance. Our hypothesis will be confirmed in the experiments of Sect.~\ref{sect:Experimental-Results}.

\subsection{Fused Geometry Augmented Images}\label{sect:Mapped-Image}
\begin{figure}[t]
\centering
\includegraphics[width=0.9\linewidth]{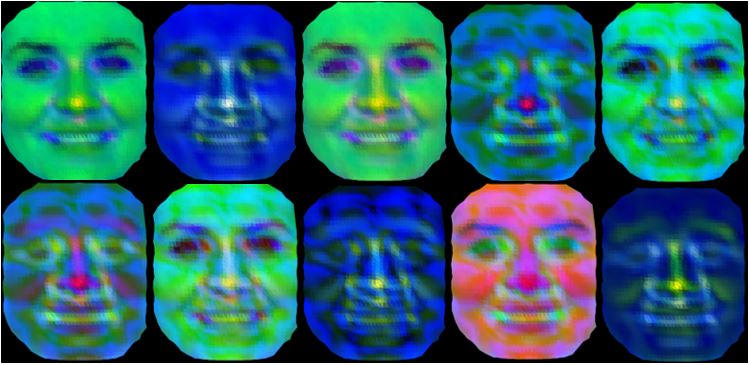}
\caption{The ten three-channel FGAIs generated for a sample face by combinatorially selecting the mapped geometric descriptors. Top row: K-GL-LD, K-H-GL, H-GL-LD, K-LD-SI, K-GL-SI. Bottom row: H-LD-SI, H-GL-SI, K-H-SI, GL-LD-SI, K-H-LD.} 
\label{fig:fused}
\end{figure}

After mapping the extracted local geometric descriptors onto the 2D textured images, we can encode the shape information with the help of a compact 2D matrix. Since we extract four local geometric descriptors, the shape information is represented by their corresponding three 2D GAIs, to which we add the gray level image.
We propose to fuse the shape information in terms of multiple descriptors by combinatorially selecting three descriptors at once. This results in ten three-channel images ($C_{3}^5$) that we call \textit{fused geometry augmented images} (FGAIs).
Each FGAI realizes a sort of early fusion between the descriptors. For example,  an FGAI can be a combination of three GAIs obtained, respectively, from Gaussian curvature (K), shape-index (SI), and gray-level (GL); thus, we can indicate this specific combination as K-SI-GL.  \NW{While we have no evidence about the plausibility of this hypothesis,  we assume  the permutations of this combination across the three channels to be  equivalent (\emph{i.e.}, K-SI-GL is equivalent to GL-SI-K, SI-K-GL, SI-GL-K, K-GL-SI, GL-K-SI).  Also, though we do not experimented it  in this present  work, we think that the other permutations can be utilized as a data augmentation technique.} 
Following this early feature fusion scheme, the FGAIs for a sample face are visualized in Fig.~\ref{fig:fused}.

\NW{If we consider the GAIs as our input data, the proposed FGAIs constitute a data-fusion scheme, as opposed to feature-level schemes, where the fusion operates on the features derived from the input images. Here, one can envisage dedicating a CNN network acting as feature extractor to each GAI type, and the derived features will be fed at fusion layer which output goes to a fully connected layer followed by an SVM or softmax classifier. This feature-level fusion architecture was adopted by Li et al.~\cite{lin2017}. We believe that our data-fusion approach will be more advantageous for three main reasons: \emph{(i)} low-level fusion performs better than higher-level counterparts when it comes to biometry applications~\cite{ross2003}; \emph{(ii)} Our data-fusion allows us to utilize pre-trained powerful architectures in a transfer-learning mode. Avoiding thus the time demanding training of CNNs created from scratch, while allowing us to gain from the effectiveness and the generality of learned features~\cite{sharif2014cnn};  \emph{(iii)} This proposed fusion scheme naturally brings-in the effect of data augmentation, which has proved its effectiveness in numerous deep learning tasks, especially where limited training data are available.}

Based on the above, we have opted to employ the FGAIs in a transfer-learning approach using two standard architectures, the AlexNet and Vgg-vd16. We propose to explore first these architectures in a reuse mode rather than the fine-tuning mode, that is, we will employ features extracted from these architectures as a starting point to the subsequent classification task. This option ensures a full relive of any training that the fine-tuning would need. 

\textbf{AlexNet} -- This network~\cite{krizhevsky2012imagenet} contains five alternating convolution and sub-sampling layers followed by Rectified Linear Units (ReLU). The network has three fully connected layers. AlexNet provided a breakthrough on the ImageNet challenge, and serves a solid baseline. We used AlexNet as a feature representation network by extracting features from two convolution layers (conv4 and conv5), one pooling layer (Pool5), and two fully connect layers (FC6 and FC7).

\textbf{Vgg-vd16} -- Compared with AlexNet, Vgg-vd16~\cite{Parkhi:2015} has far more number of trainable layers (16, with the last 3 being fully connected). The large depth of the network was possible because of smaller filter-sizes ($3\times3$). Vgg-vd16 achieved significant performance gain on the ImageNet dataset, and has shown its effectiveness on a number of transfer learning tasks. We utilized three convolution layers (conv$5_1$, conv$5_2$ and conv$5_3$) and two fully connected layers (FC6 and FC7).

A performance comparison of AlexNet and Vgg-vd16 is presented in our experimental evaluations (see Sect.~\ref{sect:Experimental-Results}). For both of them, the pre-trained weights have been used to extract features from our GAIs and FGAIs.

\subsection{Classification}\label{sect:Classification}
For each  facial expression (FE) / action unit (AU) class, we learn a discriminative model. For this purpose, we train a simple one-vs.-rest binary SVM classifier. Specifically, to learn the model parameters of one FE/AU, we consider feature encoding  for one FE/AU as the positive class, whereas the encodings of the remaining  FEs/AUs are considered as the negative class. A binary SVM is then trained to learn the hyperplane which optimally discriminates the two classes:
\begin{equation}
\label{l2l2}
\min_\mathbf{w} \; \; \frac{1}{2}\mathbf{w}^T\mathbf{w} + C \sum_t \left ( \max\left ( 0,1-{\ell_t}\mathbf{w}^T\mathbf{x}_t \right ) \right )^2 \; ,
\end{equation}

\noindent
where $\ell_t=\{1,-1\}$, $\mathbf{x}$ represents the feature vector, $\mathbf{w}$ is the vector defining the parameters of the separation hyper-plane, and $C$ corresponds to the penalty parameter. This last controls the trade-off between the speed of training and the number of support vectors.
Following this procedure, we learn a set of model parameters $\mathbf{w}_i$: $ i = 1...m$, where \textit{m} is equal to $6$ and $24$, for the FEs and AUs, respectively.
\section{Experimental Results}\label{sect:Experimental-Results}
The effectiveness of the proposed approach is demonstrated through extensive experiments in two tasks: facial expression recognition and action unit (AU) detection. We demonstrate that our proposed features are quite generic and can work simultaneously for these tasks. Experiments have been conducted with two publicly available datasets, namely, the Bosphorus 3D face database~\cite{Savran2008} and the Binghamton University 4D Facial Expression database (BU-4DFE)~\cite{yin2008}.

\subsection{Bosphorus dataset}
\begin{figure}[t]
\centering
\includegraphics[width=\linewidth, height=2.5cm]{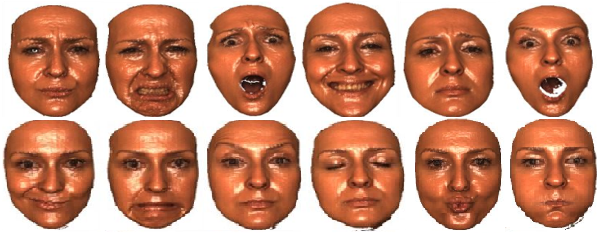}
\caption{Samples from the Bosphorus dataset. Top row: the six facial expressions, respectively, \emph{anger}, \emph{disgust}, \emph{fear}, \emph{happiness}, \emph{sadness}, \emph{surprise}.
Bottom row: six AUs out of the 24: \emph{right lip corner puller}, \emph{lip stretcher}, \emph{outer brow raiser}, \emph{eyes closed}, \emph{lip suck}, and \emph{cheek puff}.}
\label{fig:bos-sample}
\end{figure}

\NW{The Bosphorus database~\cite{Savran2008}, contains $4,666$ 3D face scans belonging to $105$ subjects. Each scan is labeled with locations of facial landmarks and presence of facial AUs and emotions (one of the six discrete facial expressions). The database is challenging due to the presence of large-scale self-occlusions and head rotations. Further, the subjects in the database exhibit diversity in terms of gender, ethnicity and age.
\NW{For facial expression recognition, $845$ scans of $65$ subjects exhibit six discrete expressions including \textit{anger}, \textit{happy}, \textit{sad}, \textit{surprise}, \textit{disgust} and \textit{fear}. For AU detection, a total of $3,838$ scans across the 105 subjects show 24 AUs.}
Samples from the Bosphorus dataset are depicted in Fig.~\ref{fig:bos-sample}.}

\NW{For facial expression recognition, we evaluate the performance in terms of accuracy, \textit{i.e.}, the ratio of correctly classified samples.
For AU detection, the area under the ROC curve (AuC) is used as a metric. Specifically, for each of the 24 AUs, we consider a one-vs.-rest classification approach and compute AuC. Other performance metrics such as Recall, Precision, F1-score, and Sensitivity could have been used for evaluation purposes as well. The choice of these two performance metrics is motivated by the need to make our results directly comparable with the reported state-of-the-art methods in the literature.}
\subsubsection{Ablative Analysis}\label{sec:ablative}
We conducted an ablative analysis experimentation that aims to investigate the performance with respect to: \emph{(i)} the discrimination capacity of the features corresponding to the different FGAIs across the different network layers; \emph{(ii)} the CNN layers employed as output features; \emph{(iii)} down-sampled data versus original data; \emph{(iv)} comparison between AlexNet and Vgg-vd16; and \emph{(v)} the early fusion scheme.

To compare the discrimination capacity of the features corresponding to the different FGAIs in the network architecture (point \emph{(i)} above), we used a Fisher's linear discriminant-like criterion. Given a feature vector $\mathcal{V}$ of length $n_\mathcal{V}$ derived from a layer $L$ in the trained CNN, for a given FGAI, we define the discrimination power of the feature $\xi^{L}_r$, $r = 1:n_\mathcal{V}$, as:
\begin{align}
\label{equ:descrit}
\mathcal{J}(\xi^{L}_r) & = \sum_{i=1}^{N_E} \sum_{j=i+1}^{N_E} \frac{1}{2}(\mu_i(\xi^{L}_r) - \mu_j(\xi^{L}_r))^2( \frac{1}{\sigma_{i}(\xi^{L}_r)^2} + \frac{1}{\sigma_{j}(\xi^{L}_r)^2} ) \\ \nonumber
& +\frac{1}{2}( \frac{\sigma_{i}(\xi^{L}_r)^2}{\sigma_{j}(\xi^{L}_r)^2} + \frac{\sigma_{j}(\xi^{L}_r)^2}{\sigma_{i}(\xi^{L}_r)^2} - 2) \; ,
\end{align}

\noindent
where $N_E$ is the number of facial expressions (\textit{i.e.}, 6) $(\mu_i,\mu_j)$, and $(\sigma_i,\sigma_j)$ are the mean and the standard deviation of the feature $\xi^{L}_r$ computed for a pair of facial expression classes $i$ and $j$, respectively.
The larger is the criterion $\mathcal{J}$, the higher is the discrimination capacity of the feature $\xi^{L}_r$. For a given layer in the network and for each facial expression class, we compute all the output feature vector samples corresponding to a specific FGAI. Then, for each element, (\textit{e.g.}, feature $\xi^{L}_r$) in the vector, we compute the criterion $J$ as described in Eq.~\eqref{equ:descrit}.
The different criteria computed for each feature are then ranked and displayed together for the different layers so that they can be visually compared. Note that for the AlexNet, a number of elements in the feature vectors are expected to have a zero-value because of the large sparsity of the weights in the network layers~\cite{lorieul2016}. Therefore, we detect and remove these zero-valued features from each feature vector before computing the criterion $J$.
\begin{figure}[t]
\centering
\begin{minipage}[h]{0.49\linewidth}
\includegraphics[width=\linewidth]{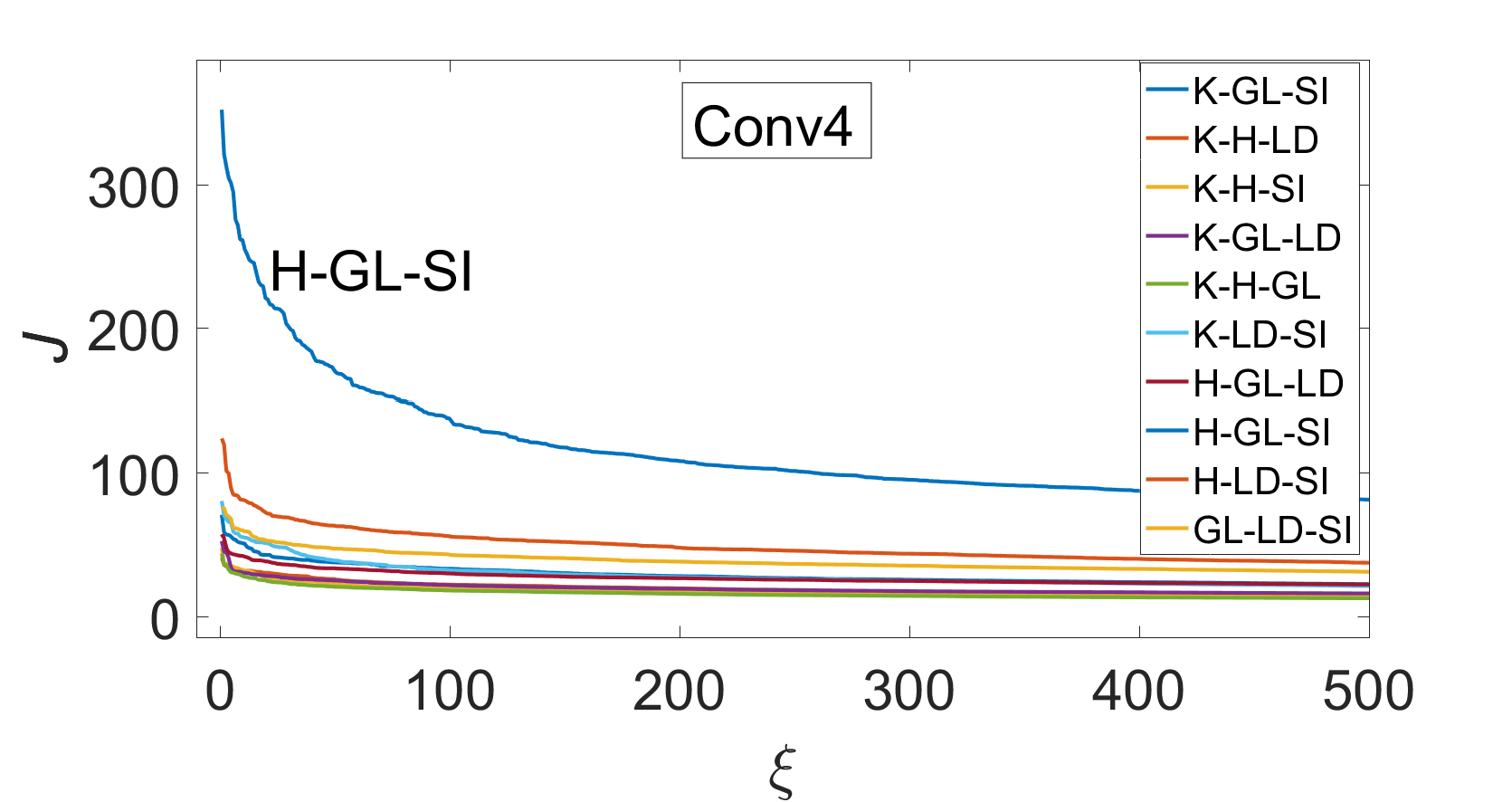}
\includegraphics[width=\linewidth]{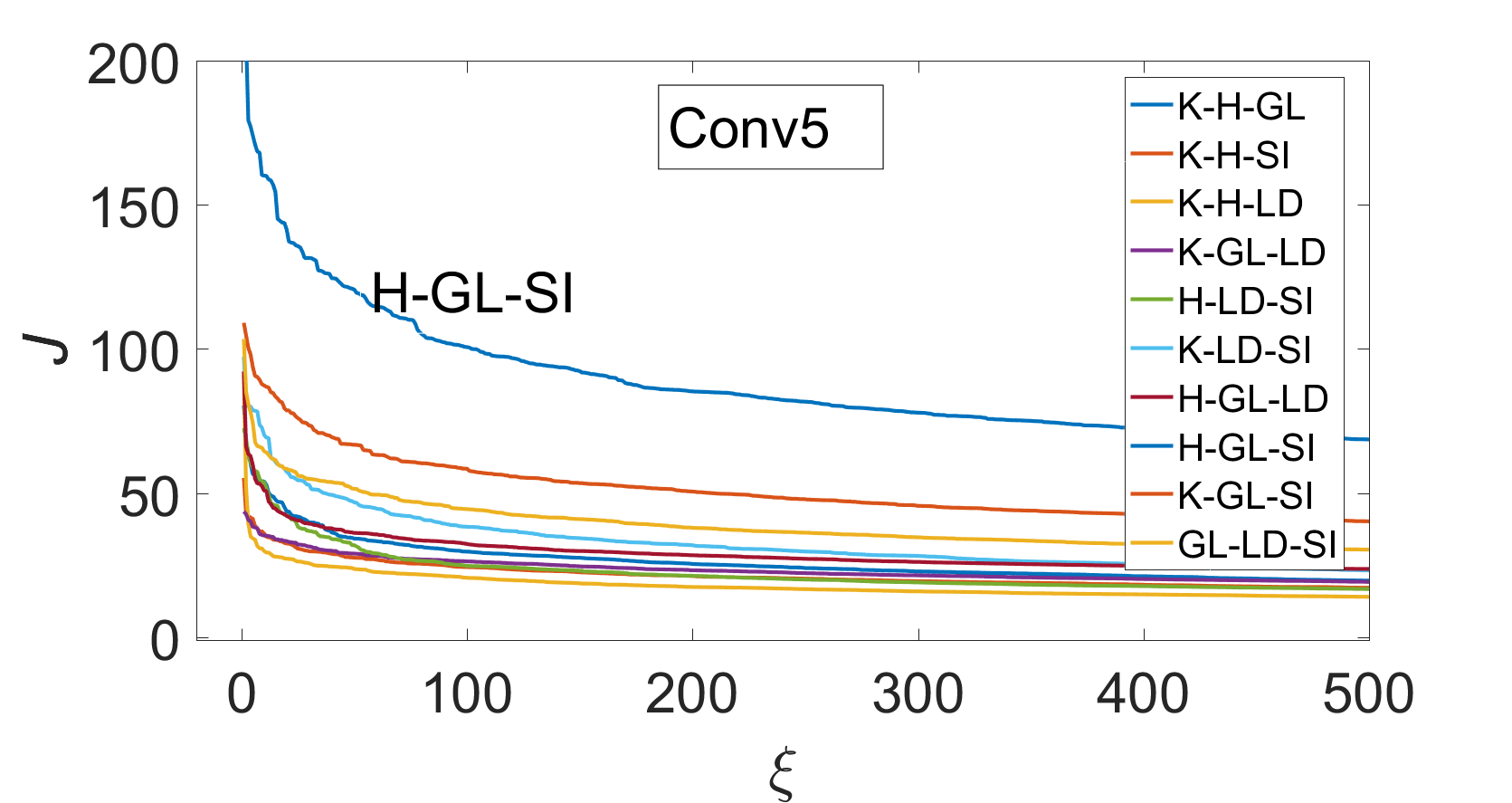}
\includegraphics[width=\linewidth]{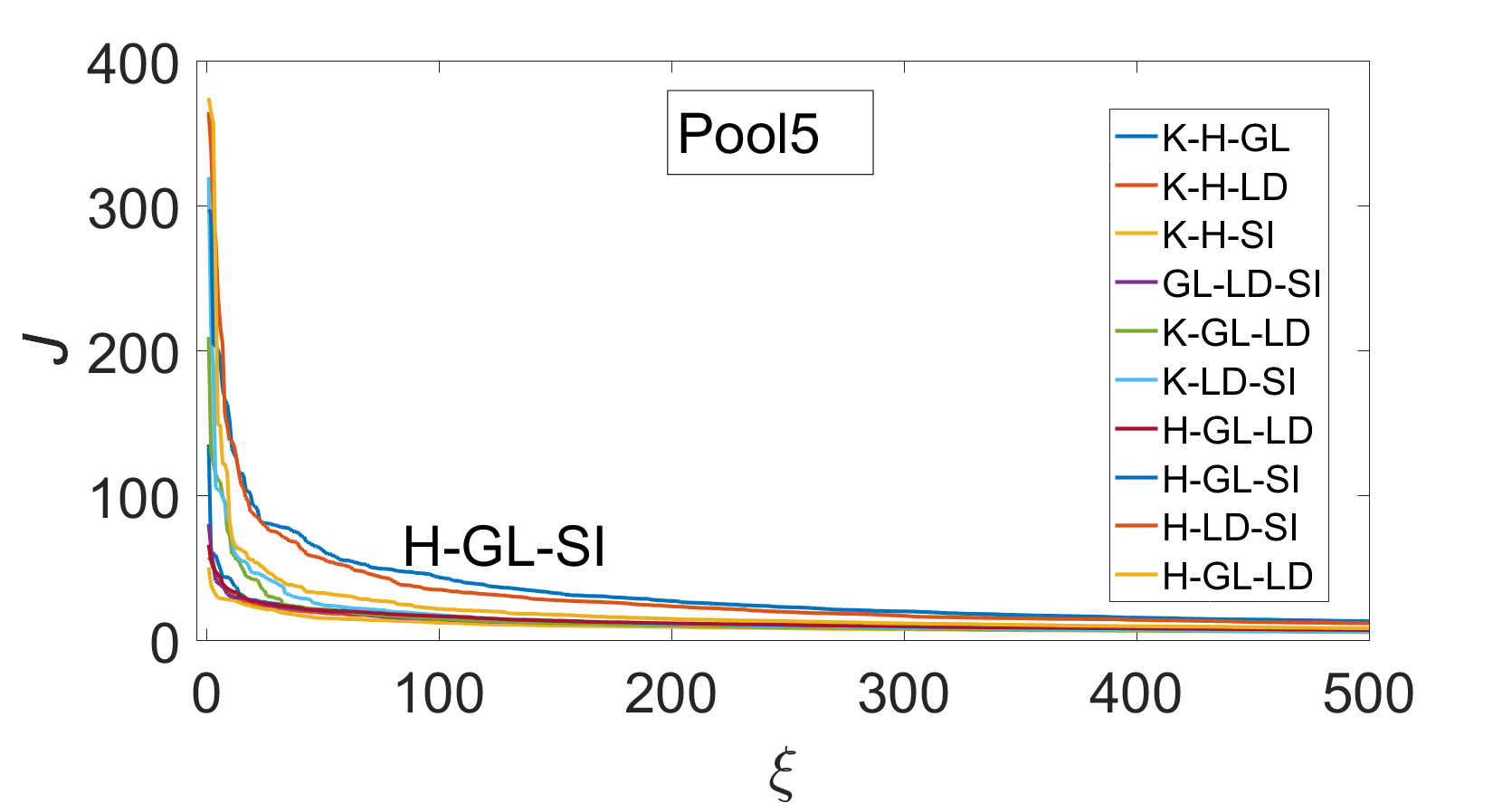}
\end{minipage}
\begin{minipage}[h]{0.49\linewidth}
\includegraphics[width=\linewidth]{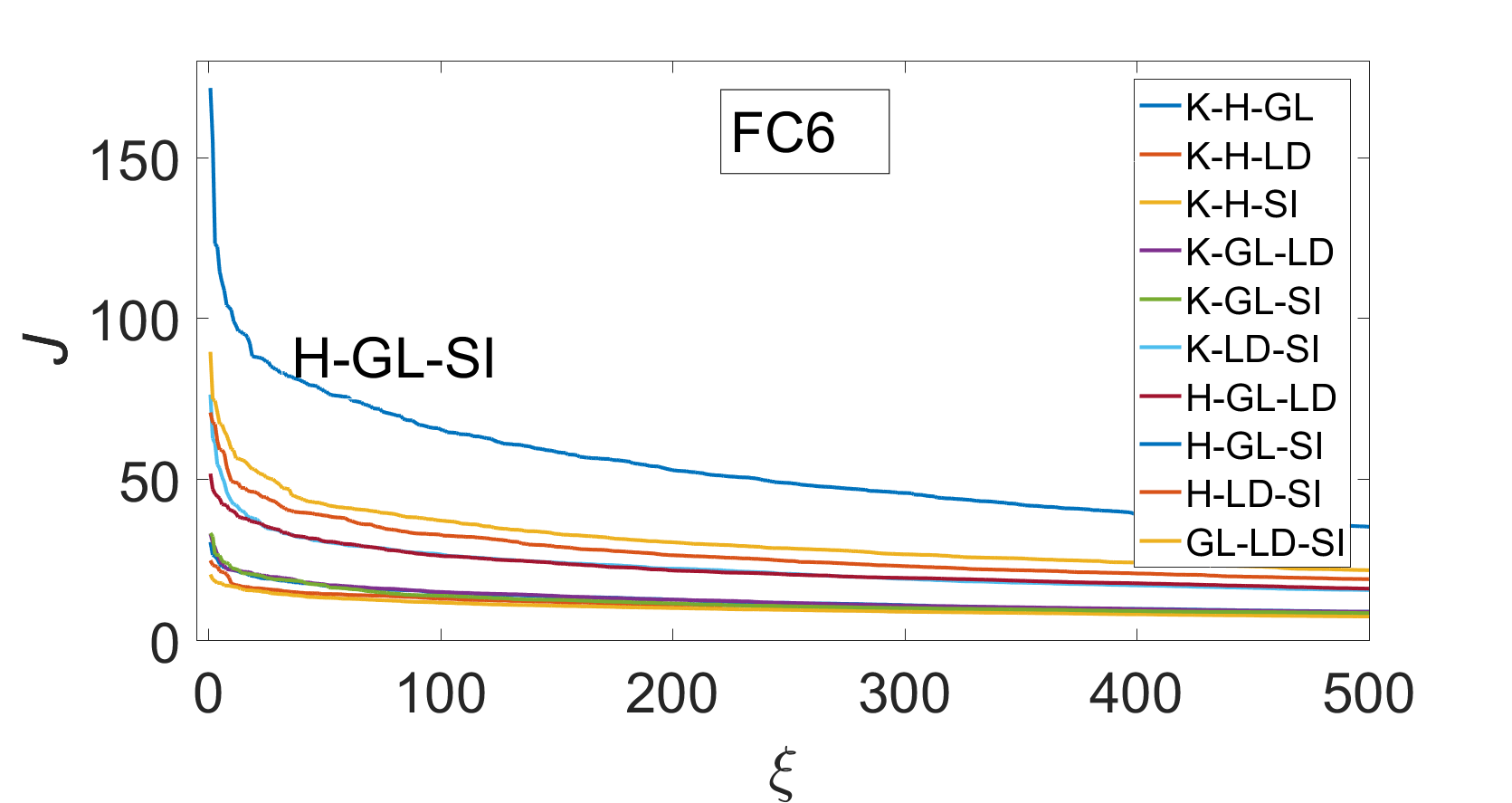}
\includegraphics[width=\linewidth]{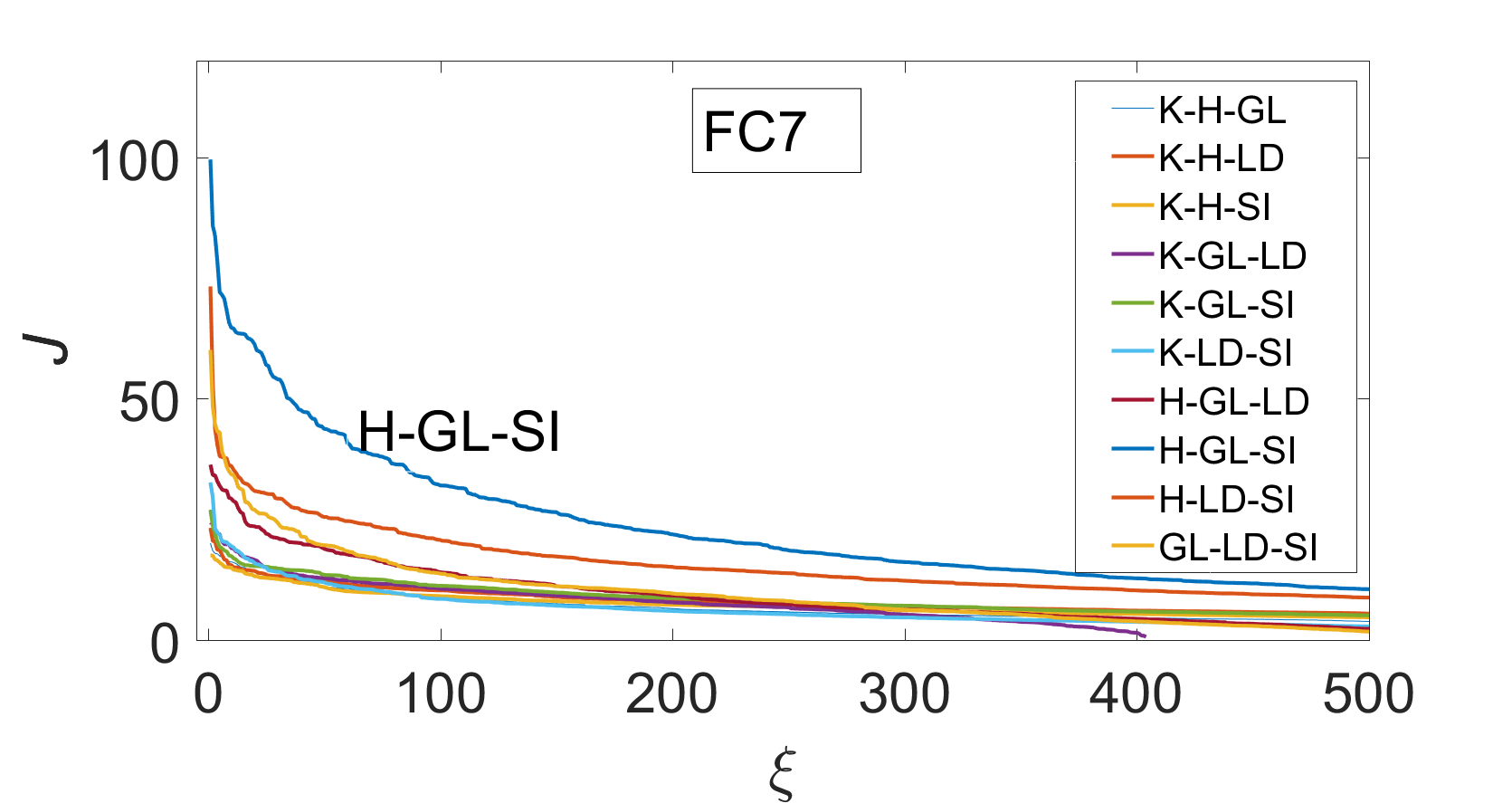}
\includegraphics[width=\linewidth]{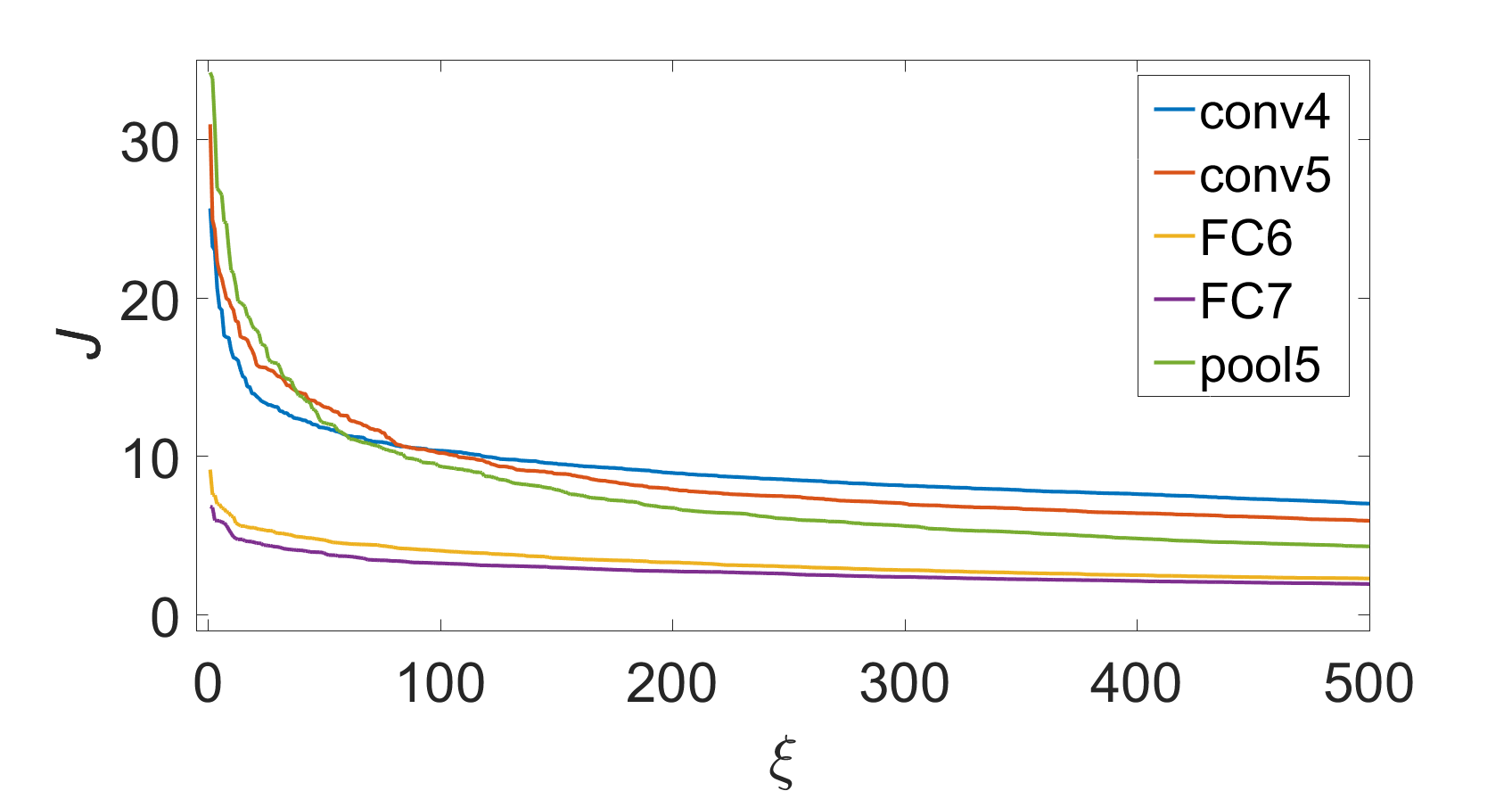}
\end{minipage}
\caption{Bosphorus dataset: Ranked $\mathcal{J}$ of Eq.~\eqref{equ:descrit} plotted for the top $500$ features for each FGAI in the AlexNet layers.
\SB{Looking to the plots top-to-bottom and left-to-right, the first five compare the ten FGAI at Conv4, Conv5, Pool5, FC6 and FC5 network layers; the last plot compares the different layers to each other in the same plot for all the FGAI.}}
\label{fig:descri_power_alexnet}
\end{figure}
Figure~\ref{fig:descri_power_alexnet} depicts the criteria $J$ for the top $500$ features corresponding to the ten FGAIs across the AlexNet layers.
The first five plots (top-to-bottom and left-to-right) show that the FGAI for the combination H-GL-SI looks having the largest discriminative capacity neatly above the others, particularly at Conv4 and Conv5.
\SB{The other plots show some disparity for the other FGAIs, thus not allowing a conclusive assessment as for the H-GL-SI.
In the last plot in the figure (third row, second column)}, we report a test variant aiming to assess the discrimination power of the features layer-wise (\emph{i.e.}, for each of Conv4, Conv5, Pool5, FC6 and FC7). In particular, we computed the criterion $\mathcal{J}$ for each feature in these layers, considering all the FGAI outputs when computing the means and the standard deviations in Eq.~\eqref{equ:descrit}.
Results clearly show that the criterion $\mathcal{J}$ keeps neatly higher in the Conv4, Conv5 and Pool5 layers compared to FC6 and FC7 across all the features, thus reflecting a higher discrimination capacity of the former layers.

\SB{The $J$ test above accounts for the general discriminative power of the different FGAI at different layers; to further investigate the effectiveness of the features,} we conducted a series of tests assessing the performance of each FGAI for the facial expression recognition across the different network layers of AlexNet and Vgg-vd16. We considered the original version of the face scans (\emph{i.e.}, without any down-sampling).   We adopted the standard 10-fold cross validation over 60 randomly selected subjects,  by partitioning the subjects into 10 sets, and deriving, in each round,  the testing test  from one fold, and the training set from the other 9 folds. 
Table~\ref{tab:combBos} reports the classification rate obtained for AlexNet~(a) and Vgg-vd16~(b). Looking at the first and the second score in each column, marked in bold and blue, respectively, we can notice that K-GL-SI and H-GL-SI form the best combinations for AlexNet. It is also noticeable that the pair GL-SI is present in both of them. For the Vgg-vd16, we can observe that H-GL-SI and H-LD-SI seem representing the best combinations.
We also notice the low rate obtained with FC6 and FC7 in both the networks confirming the discriminative power analysis reported earlier in Fig.~\ref{fig:descri_power_alexnet}.
Referring to the overall findings in the discrimination analysis and the facial expressions, we selected the H-GL-SI as the best FGAI candidate for the rest of the experimentation.
\begin{table}[t]
\caption{Bosphorus dataset: Classification rate using the ten different FGAIs for AlexNet~(a) and Vgg-vd16~(b)}
\label{tab:combBos}
\begin{minipage}[h]{0.05\linewidth}
(A)
\end{minipage}
\begin{minipage}[h]{0.43\linewidth}
\begin{tabular}{lccccc}
\toprule
{FGAI} & {Conv4} & {Conv5} & {Pool5} & {FC6} & {FC7}  \\
\midrule
K-H-GL & {89.48}           & {89.0} 				& {89.56} 			& {44.44} 			& {42.22}  \\
K-H-LD  & {91.67}            & {85.67} 				& {90.67} 			& {42.78} 			& {45.67} \\
K-H-SI  & {95.7}             & {90.11} 				& {91.22} 			& {41.78} 			& {25.56} \\
K-GL-LD & {94.6}             & {92.33} 				& {92.53} 			& {47.78} 			& \bc{{50.0}} \\
K-GL-SI & \bc{97.53}         & \textbf{{96.78}} 	& \bc{97.33} 	        & {52.33}           & {37.78} \\
K-LD-SI  & {96.44}           & {90.67}          	& {96.78}           & {51.78} 								& {30.0}  \\
H-GL-LD & {94.97}            & {93.44}          & \textbf{{97.89}}  & {47.89} 								& \textbf{{50.0}}  \\
H-GL-SI & \textbf{98.27}     & \bc{94.56}         		& {96.22}            & \bc{53.44} 								& {46.22} \\
H-LD-SI & {96.8}             & {89.56}         		& {96.78}            & \textbf{{54.56}}								& {41.22} \\
GL-LD-SI & {96.07}           & {89.56}         		& {90.67}            & {47.33} 							  & {22.22} \\
\bottomrule
\textbf{Mean}     &    95.153           & 91.168  & 93.965  & 48.411   & 39.089  \\
\bottomrule
\end{tabular}
\end{minipage}

\vspace{0.2cm}

\begin{minipage}[h]{0.05\linewidth}
(B)
\end{minipage}
\begin{minipage}[h]{0.43\linewidth}
\begin{tabular}{lccccc}
\toprule
{FGAI} & {Conv$5_1$} & {Conv$5_2$} & {Conv$5_3$} & {FC6} & {FC7}  \\
\midrule
K-H-GL & {94.59}       & {89.2} 			& {88.44} 		  & {64.37} 		& {56.22}  \\
K-H-LD  & {92.89}        & {88.31} 			& {85.67} 		  & {\textbf{66.03}} 		& \bc{63.44} \\
K-H-SI  & {93.44}        & {89.4} 			& {87.89} 		  & {55.04} 		& {57.33} \\
K-GL-LD & {95.44}        & {93.33} 			& {89.56} 		  & {61.27} 		& {62.89} \\
K-GL-SI & {97.44}        & {95.53} 			& {\textbf{94.0}} & {62.85} 		& {61.78} \\
K-LD-SI  & {97.54}       & {96.10} 			& {89.56} 		  & {61.38} 		& {60.11}  \\
H-GL-LD & {97.89}        & \bc{93.7} 		& {91.78} 		  & {60.54} 		& {\textbf{67.33}}  \\
H-GL-SI & \textbf{98.16} & {92.5} 			& {91.22} 		  & {\bc{65.78}} & {57.89} \\
H-LD-SI & \bc{98.03}     & {\textbf{96.12}} & \bc{93.44} 	  & {58.85} 		& {59.56} \\
GL-LD-SI & {96.78}       & {89.76} 			& {85.67} 		  & {51.38} 		& {55.11} \\
\bottomrule
\textbf{Mean}     &     96.22    &  92.395      &  89.723 &  60.549  & 60.166 \\
\bottomrule
\end{tabular}
\end{minipage}
\end{table}

\begin{figure}[ht]
\centering
\includegraphics[width=0.5\linewidth]{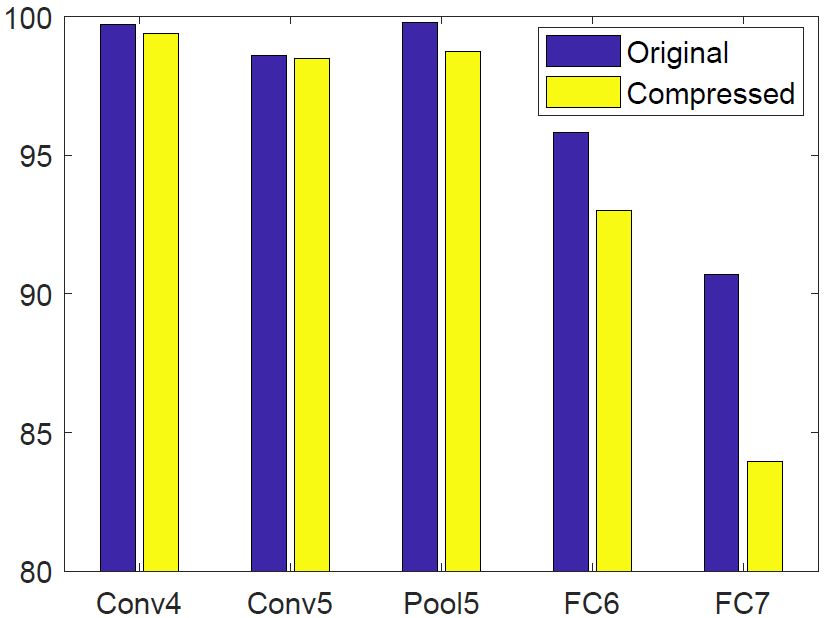}
\includegraphics[width=0.48\linewidth]{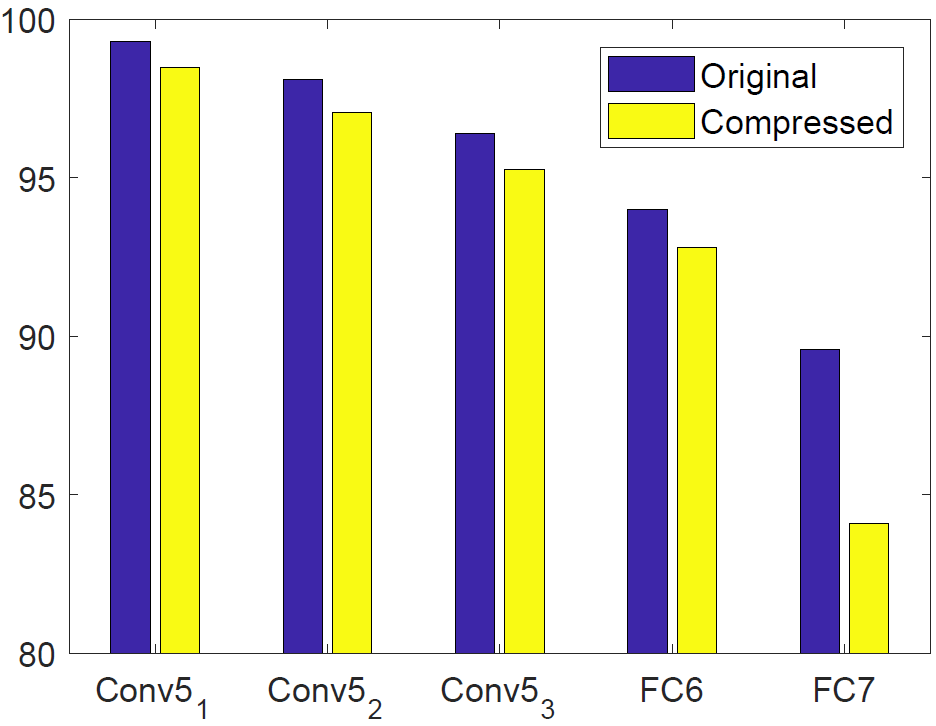}
\caption{Bosphorus dataset, AU detection: AuC values comparison for features extracted at different layers of  AlexNet (left), and Vgg-vd16 (right). Comparison between the results obtained with the original (blue bar) and compressed, \emph{i.e.}, down-sampled (yellow bar) data is also shown.}
\label{fig:layers}
\end{figure}

For the aspects \emph{(ii)}, \emph{(iii)}, and \emph{(iv)}, we conducted an experimentation on AUs classification with AlexNet and Vgg-vd16. We used features computed with the FGAI (H-GL-SI) from both the original and the down-sampled data, in a $5$-fold cross validation. Results of the analysis are summarized in Fig.~\ref{fig:layers}~(a)-(b) showing the average AuC for AU detection.

For AlexNet, we notice that for the original data, the highest score is obtained with the max pooling layer Pool5 ($99.79\%$) followed by the convolutional layers Conv4 ($99.71\%$) and Conv5 ($98.6\%$); then, we observed a drop in the recognition rate in the fully connected layers, particularly noticeable for FC7 ($90.7\%$). This decrease, in concordance with the discriminative analysis of Fig~\ref{fig:descri_power_alexnet}, can be explained by the fact that the deeper and final layers in the model are more ImageNet class-specific.
A similar behavior is observed for Vgg-vd16 that shows the best recognition rate for the Conv4 layer ($99.3\%$).

With respect to the proposed down-sampling scheme of the data, we notice a same moderate decrease pattern in both AlexNet and Vgg-vd16. Second, in AlexNet, the drop in the recognition rate is more noticeable in the fully connected layers, particularly for FC7 (10\%), while it is quite minor in the others (less than 2\%). The same observation can be mentioned for Vgg-vd16, for the first four layers, while the last layer (FC7) shows a relatively larger decrease of $5\%$. These observations suggested us the option of employing the down-sampled 3D face scans, thus allowing us a significant reduction of the computational time. \NW{Tested on an Intel i7-5500, 2.4 GHz, 16GB RAM, and 64 bits machine, we found that computation of the GAIs runs runs 11 times faster for the compressed data version than its original counterpart. This demonstrates  the significant gain in efficiency that our  down-sampling scheme affords.} 
An overall comparison of the classification rate across the different layers in the two architectures seems to be in favor of AlexNet, noticeably at the non-fully connected layers.

Finally, we investigated the extent to which the proposed early fused representation impacts the accuracy of the classification (aspect \emph{(v)} above). To this end, we conducted a series of experiments  to evidence that the improvement in the performance in our method emanates from the proposed fused scheme rather than from the usage of the pre-trained AlexNet and Vgg-vd16 networks. We investigated this aspect with experiments on facial expression classification.
\SB{The single GAI, derived from the compressed face model and corresponding to the K, H, GL, LD, and GL descriptors, is replicated over the three-channels of the network input. The features extracted from the networks are then used as the SVM training and testing data.}
\NW{From this initial set of GAI, we generated also another augmented set using a horizontal flip, rotation, and addition of white Gaussian noise. We did this to enlarge the number of GAI samples per expression, and thus to compensate any potential SVM over-fitting effect that might compromise the single GAI performance in favor of the FGAI.} Afterwards, we performed the two classification testing in $10$-fold cross-validation for each GAI.
Table~\ref{tab:bosphorus-features} reports the best classification rate obtained with a single descriptor in each layer for AlexNet and Vgg-vd16 together with the classification rate obtained with the FGAI (H-GL-SI). As it can be clearly noticed, the significant gap between the performance brings evidence that the learning from different descriptors individually is less effective than learning from the data fused using our proposed fusion scheme.
\begin{table}[ht]
\caption{Bosphorus dataset: Classification rates obtained with single GAI versus  the FGAI:H-GL-SI. }
\label{tab:bosphorus-features}
\centering
\small
\begin{tabular}{lcccccc}
\toprule
{AlexNet}  & { K} & { H} & { GL} & { LD} & { SI}  & {H-GL-SI}  \\
\midrule
\textbf{Conv4}  &  75.1 & 81.2 & 72.3 & 82.1 &88.5  & \textbf{98.27} \\
Conv5                  & 71.5 & 76.67 & 70.4 & { 79.44} & {82.79} & {\textbf{94.56}} \\
pool5                   & 73.1 & {84.12} & 72.2 & 81.88 & 78.85 & {\textbf{96.78}} \\
FC6                       &  45.1 & {52.2} & 49.8 & 52.7 & 53.9 & {\textbf{53.56}} \\
FC7                       & 39.2 & 40.16 & 40.02 & {41.48}  & 41.77 & {\textbf{46.22}} \\
\bottomrule
\\
{Vgg-vd16} \\
\midrule
\textbf{Conv$5_1$} & 80.22 & {83.48} & 78.33 & 81.17 & 80.54  & \textbf{98.16} \\
Conv$5_2$                 & 70.56 & 71.22 & {74.85} & 72.58 & 70.93 & {\textbf{92.5}} \\
Conv$5_3$                 & 73.35  & 70.49 & 72.11 & 69.98 & 70.04  & {\textbf{91.22}} \\
FC6                               & 52.48 & 54.33 & 50.62 & {58.95} & 56.42 & {\textbf{65.78}} \\
FC7                               & 39.51 & {44.08} & 40.3 & 43.5 & 42.71 & {\textbf{57.89}} \\
\bottomrule
\end{tabular}
\end{table}
\begin{figure*}[t]
\centering
\includegraphics[width=\linewidth,height=5cm]{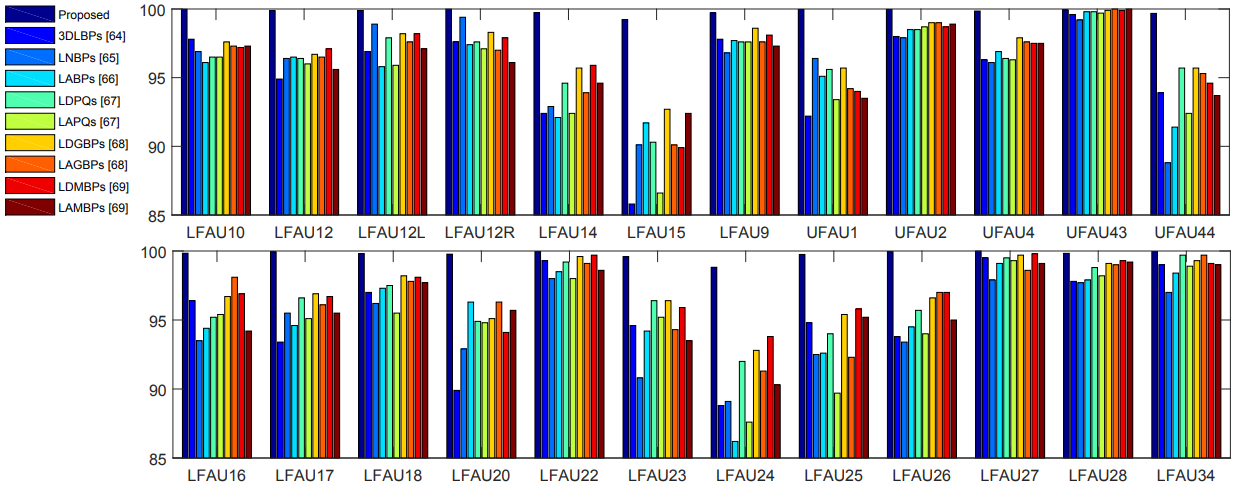}
\caption{Bosphorus dataset: AU detection measured as AuC for each of the 24 AUs. \SB{Results for 12 AUs are repotted in the upper plot, and for the remaining 12 in the lower one (please note that AUs are indicated with the abbreviation used in the Bosphorus dataset, where LF and UF stay, respectively, for lower- and upper-face, while R and L indicate the left and right part of the face).}
Comparison with the following state-of-the-art methods: 3DLBPs~\cite{Huang06}, LNBPs~\cite{Sandbach12}, LABPs~\cite{Sandbach2012}, LDPQs~\cite{Ville2008}, LAPQs~\cite{Ville2008}, LDGBPs~\cite{Zhang05}, LAGBPs~\cite{Zhang05}, LDMBPs~\cite{Yang10}, LAMBPs~\cite{Yang10}.}
\label{fig:au-results}
\end{figure*}
\subsubsection{Action Unit Classification}\label{sect:result}
\SB{In this part we experimented the extent to which our method can correctly classify each the 24 AUs.}
We compared our results with a number of existing methods, adopting the same protocol \NW{(\emph{i.e.}, $5$-fold cross validation, $3,838$ AU scans collected from $105$ subjects)}.
The state-of-the-art methods we compared to, include the 3D Local Binary Patterns (3DLBPs)~\cite{Huang06}, Local Azimuthal Binary Patterns (LABPs)~\cite{Sandbach2012}, Local Depth Phase Quantisers (LDPQs)~\cite{Ville2008}, Local Azimuthal Phase Quantisers (LAPQs)~\cite{Ville2008}, Local Depth Gabor Binary Patterns (LDGBPs)~\cite{Zhang05}, Local Azimuthal Gabor Binary Patterns (LAGBPs)~\cite{Zhang05}, Local Depth Monogenic Binary Patterns (LDMBPs) and Local Azimuthal Monogenic Binary Patterns (LAMBPs)~\cite{Yang10}.

Figure~\ref{fig:au-results} shows the AuC results for each AU individually, obtained with our proposed method, using the combination (H-GL-SI), together with the state-of-the-art methods listed above.
\SB{Computing the average AuC on all the AUs,} it can be observed that our proposed feature representation scheme achieves the highest score of $99.79\%$, outperforming the current state-of-the art AuC of $97.2\%$, which is scored by the Depth Gabor Binary Patterns (LDGBP) feature proposed in~\cite{Sandbach2012}.

\begin{table}[htb]
\caption{Bosphorus dataset: Classification rate obtained with the FGAI:H-GL-SI compared with the state-of-the-art methods.}
\label{tab:bosph:FE}
\center
\begin{tabular}{lc}
\toprule
{Method} & {Accuracy} \\
\hline
MS-LNPs~\cite{li2012a}         & {75.83} \\
GSR~\cite{Yang15}                  & {77.50} \\
iPar–CLR~\cite{LI2015}          & {79.72} \\
DF-CNN svm~\cite{lin2017} & {80.28} \\
Original-AlexNet                       & \textbf{{98.27}} \\
Compressed-AlexNet              & \textbf{{93.29}} \\
Original-Vgg-vd16                   & \textbf{{98.16}} \\
Compressed-Vgg-vd16          & \textbf{{92.38}} \\
\bottomrule
\end{tabular}
\end{table}

\subsubsection{Facial Expression Classification}
\NW{For facial expression recognition, we adopted the same experimentation protocol reported in the recent state-of-the-art methods~\cite{li2012a,Yang15,LI2015,lin2017} ($10$-cross validation, expression scans collected from $60$ subjects, randomly selected from $65$ individuals). 
Results obtained with the (H-GL-SI) are reported in Table~\ref{tab:bosph:FE} for both original and compressed face scans.
Remarkably, the proposed scheme, for both original and compression face scans, outperforms the current state-of-the-art solutions, by a significant margin of $18\%/13\%$ and $17\%/12\%$ for the learned features of the AlexNet and the Vgg-vd16, respectively.}

\subsection{BU-4DFE}
\begin{figure}[b]
\centering
\begin{minipage}[c]{0.1\linewidth}
\footnotesize{(a)}
\end{minipage}
\begin{minipage}[c]{0.87\linewidth}
\includegraphics[width=\linewidth, height=1cm]{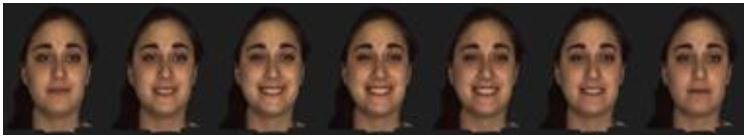} 
\end{minipage}

\begin{minipage}[c]{0.1\linewidth}
\footnotesize{(b)}
\end{minipage}
\begin{minipage}[c]{0.87\linewidth}
\includegraphics[width=\linewidth, height=1.1cm]{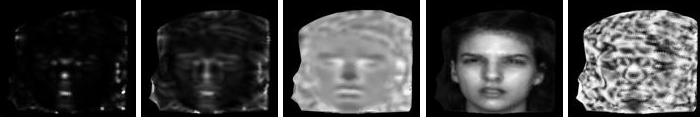}
\end{minipage}
\caption{BU-4DFE dataset: (a) Sample 3D frames selected from the \textit{neutral}, \textit{onset}, \textit{apex}, \textit{offset}, and \textit{neutral} parts of a sequence labeled as ``Happy''; (b) GAIs extracted from sample frames of one  subject with reduced resolution to $240 \times 240$. From left: Gaussian curvature (K), mean curvature (H), local depth (LD), gray level (GL) and shape index (SI).}
\label{fig:bu-sample}
\end{figure}

\NW{The BU-4DFE dataset~\cite{yin2008} contains $101$ subjects divided into $43$ males and $58$ females. Each subject is captured in six different expressions: \textit{Anger}, \textit{Disgust}, \textit{Fear}, \textit{Happy}, \textit{Sad} and \textit{Surprise}. Each video contains a sequence of 3D face scans captured at the rate of $25fps$ for a $4$ seconds duration. The expression dynamics encompasses four phases: \textit{neutral}, \textit{onset}, \textit{apex}, \textit{offset}, and \textit{neutral} (see Fig.~\ref{fig:bu-sample}.a for an example). All the experiments reported in this section have been conducted with the down-sampled version of the BU-4DFE scans (\emph{i.e.}, compression equal to 3.5)}.
Figure~\ref{fig:bu-sample}.b depicts samples of the GAI for one subject.
\subsubsection{Ablative Analysis}
\begin{figure}[htb]
\centering
\begin{minipage}[h]{0.49\linewidth}
\includegraphics[width=\linewidth]{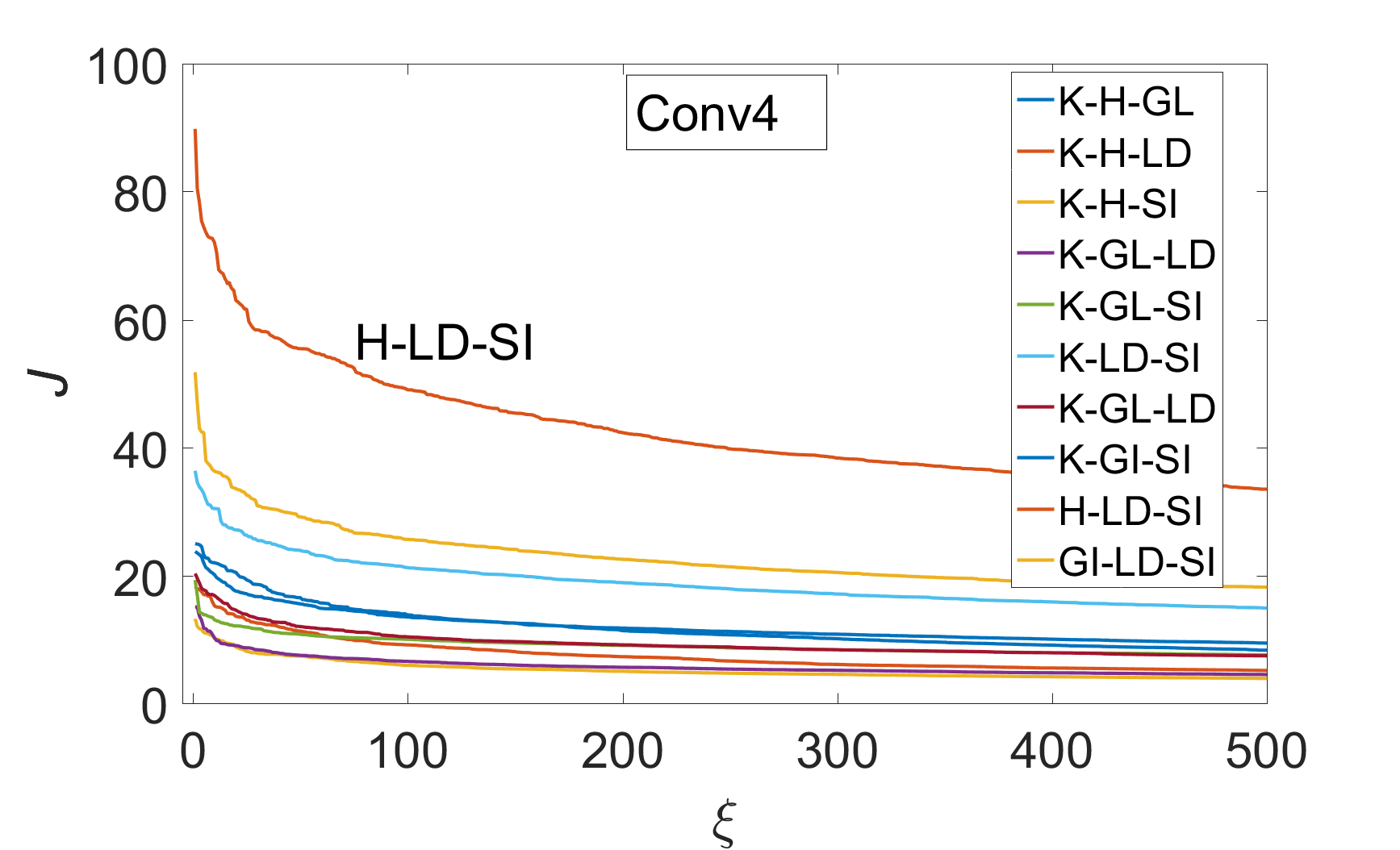}
\includegraphics[width=\linewidth]{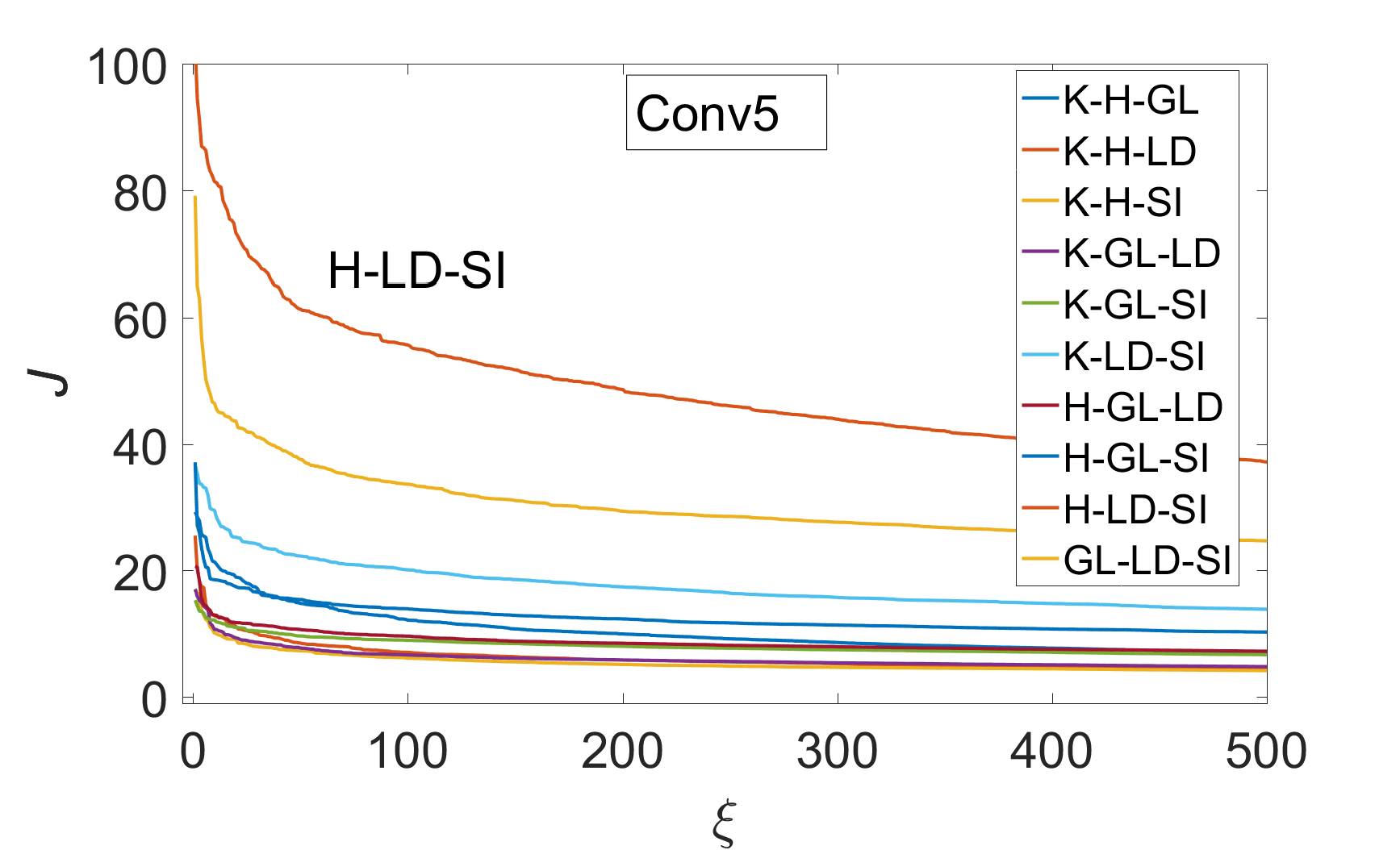}
\includegraphics[width=\linewidth]{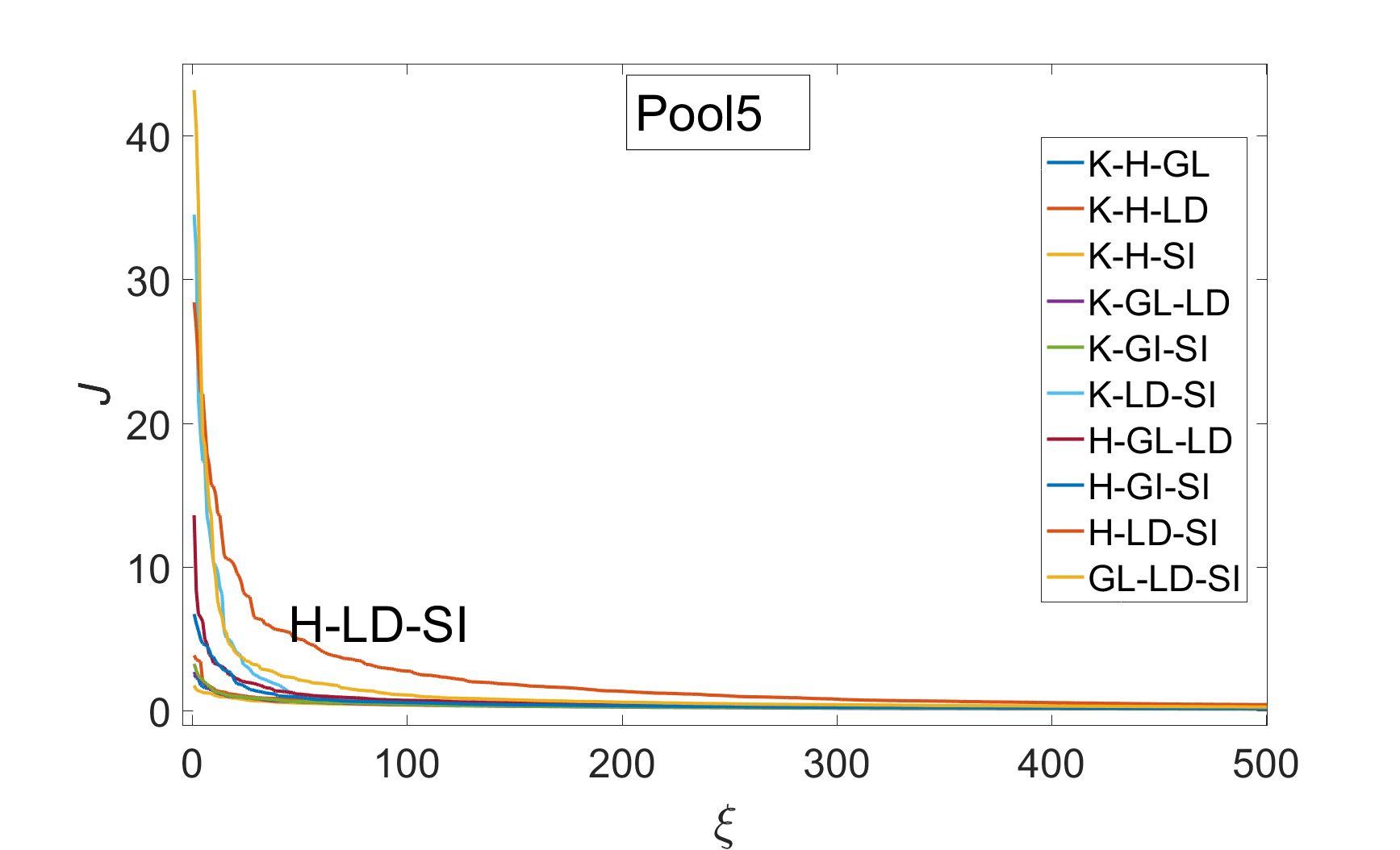}
\end{minipage}
\begin{minipage}[h]{0.49\linewidth}
\includegraphics[width=\linewidth]{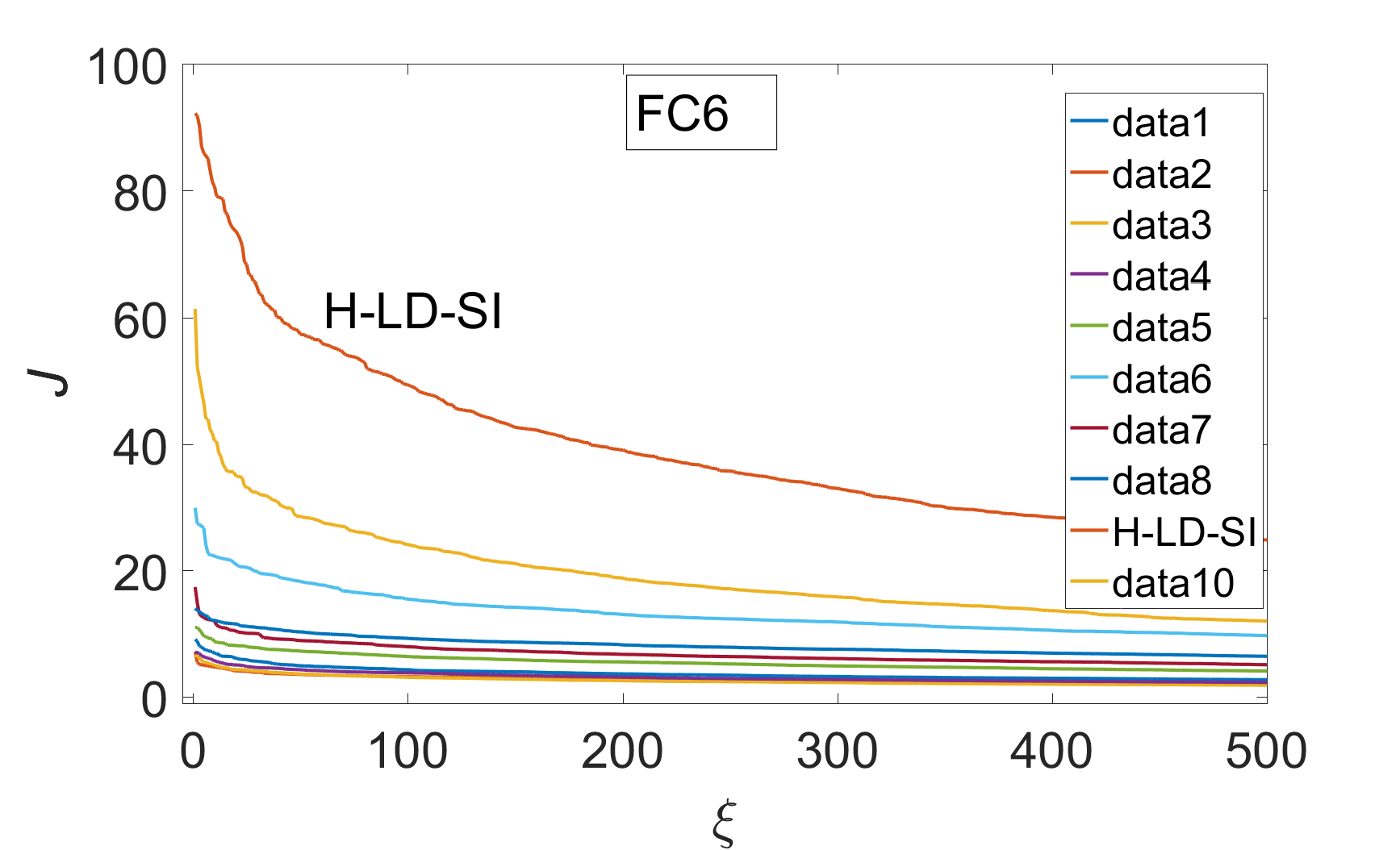}
\includegraphics[width=\linewidth]{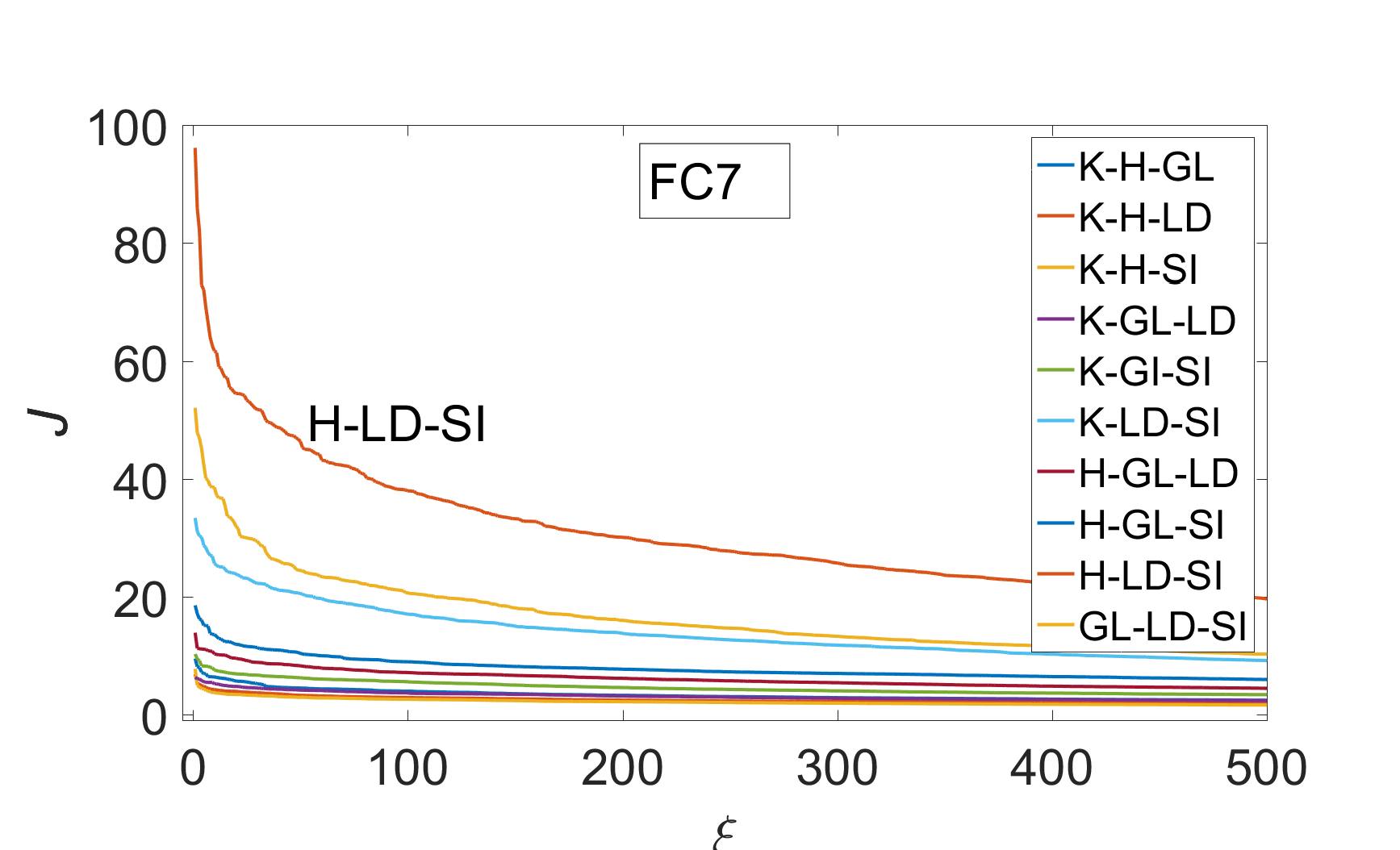}
\includegraphics[width=\linewidth]{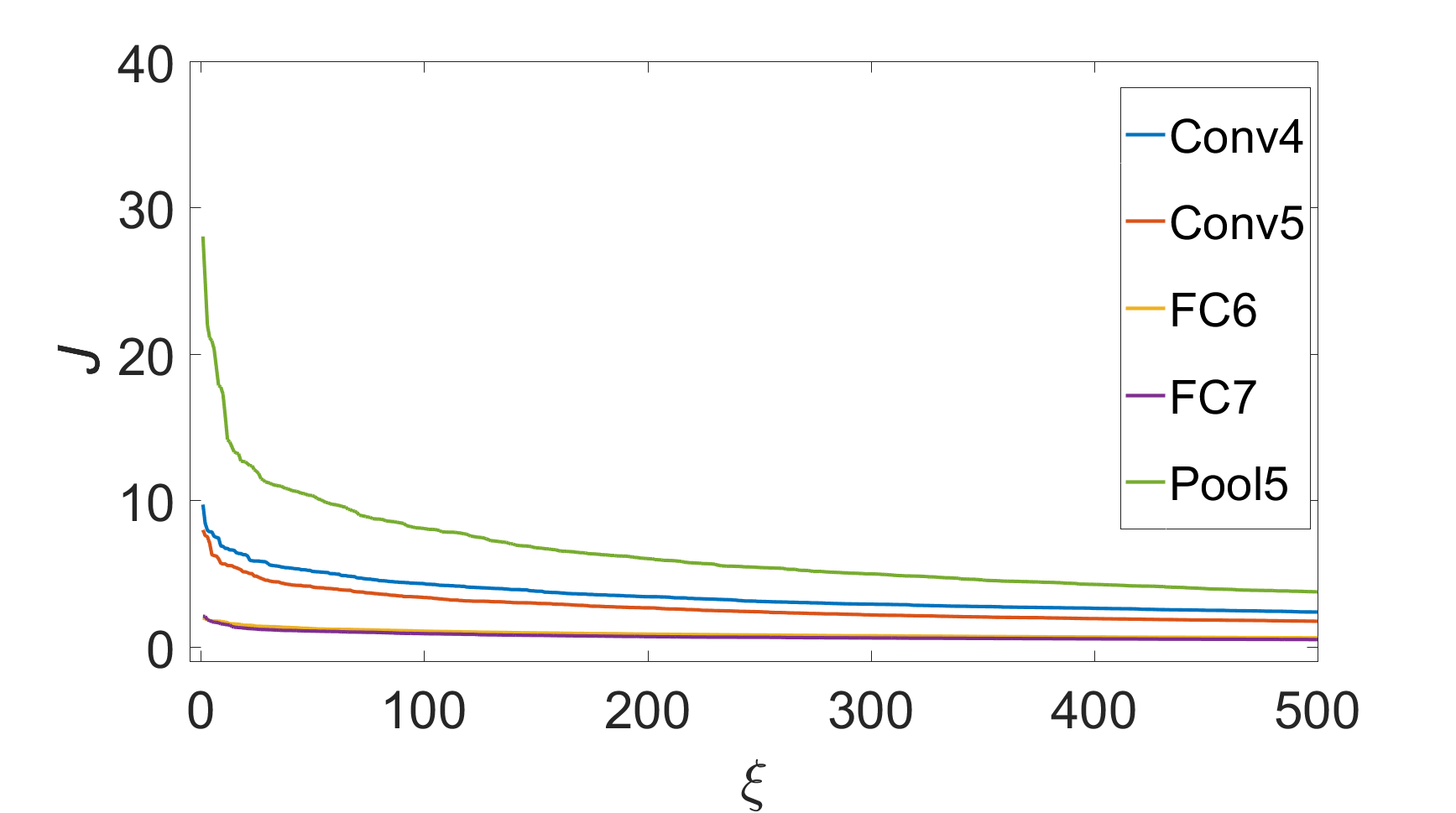}
\end{minipage}
\caption{BU-4DFE dataset: Ranked $\mathcal{J}$ of Eq.~\eqref{equ:descrit} plotted for the top $500$ features for each FGAI in the  AlexNet layers.}
\label{fig:descri_power_alexnet_bu4d}
\end{figure}

We conducted  the same experiments described for the Bosphorus dataset (see Sect.~\ref{sec:ablative}) investigating the performance of the different FGAIs, the network layers, and the impact of the early fusion.
Figure~\ref{fig:descri_power_alexnet_bu4d} depicts the criteria $J$ for the top $500$ features corresponding to the ten FGAIs across the AlexNet layers. Here we found that the FGAI with the H-LD-SI combination exhibited the highest discrimination power. Layer-wise, as with the Bosphorus dataset, Conv4, Conv5 and Pool5 showed better discrimination capacity compared to FC6 and FC7.

\NW{Results for the experiment assessing the performance of the FGAIs individually for the facial expression classification obtained with AlexNet is reported in Table~\ref{tab:combBU}-(A). Here, we considered the six expressions plus the neutral expression, which data can be derived from the video sequences, and proceeded in a 5-fold cross validation.
We can observe that all the FGAIs produce close and high classification rate at Conv4, the same can be said for Pool5 and Conv5, though with slightly lower performance. The classification rate drops consecutively for the FC6 and the FC7 layer.
Overall, the FGAIs with the H-LD-SI combination achieved the best performance across the different layers, which concord with the discrimination analysis previously reported in Fig.~\ref{fig:descri_power_alexnet_bu4d}. 
The classification rate obtained with the Vgg-vd16(Table~\ref{tab:combBU}-(B)) indicates that the FGAI:H-LD-SI has the best performance, confirming the results found in the discrimination analysis of Fig.~\ref{fig:descri_power_alexnet_bu4d}. Based on these findings, we considered H-LD-SI as the best FGAI, and utilized it in the following experiments.}
\begin{table}[htb]
\caption{BU-4DFE dataset: Classification rate using the ten different FGAIs for AlexNet~(a) and Vgg-vd16~(b).}\label{tab:combBU}
\begin{minipage}[h]{0.05\linewidth}
(A)
\end{minipage}
\begin{minipage}[h]{0.43\linewidth}
\begin{tabular}{lccccc}
\toprule
{FGAI}     & {Conv4}            & {Conv5}            & {Pool5}       & {FC6}               & {FC7}  \\
\midrule
K-H-GL   & {89.98}                 &{\textbf{87.84}}       & {{89.84}}    & {71.2}            & {\textbf{54.05}}  \\
K-H-LD   & {89.95}			         & {85.02}	              & \bc{89.95}          & {67.02}	        & {30.16} \\
K-H-SI    & {88.27}			         & {81.66}	              & {88.02}	            & {62.34}	        & {18.48} \\
K-GL-LD & {\bc{90.09}}           & {86.84}                & {89.8} 	            & {70.77}        	& {40.98} \\
K-GL-SI  & {88.34} 			         & {85.88}       	      & {87.84}          	& {69.66}       	& {27.84} \\
K-LD-SI  & {86.88} 			         & {83.63} 	              & {86.84} 			& {66.77} 	        & {25.66}  \\
H-GL-LD & \bc{90.09} 	        & \bc{87.77} 	          & {89.73} 		    & {{71.84}}	& {42.23}  \\
H-GL-SI  & {88.48} 			     	& {86.27} 	              & {87.7} 				& \bc{71.34}        & {27.05} \\
H-LD-SI  & \textbf{91.34} 			& {86.8} 	              & \textbf{90.02} 		    & \textbf{72.3} 	        & \bc{48.09} \\
GL-LD-SI & {88.56} 			     	& {82.48} 	              & {86.48} 	        & {59.52}       	& {24.2} \\
\bottomrule
\textbf{Mean}     &	90.03	& 	87.34	    & 89.89 &    71.27&  33.87  \\
\bottomrule
\end{tabular}
\end{minipage}

\vspace{0.2cm}
\begin{minipage}[h]{0.05\linewidth}
(B)
\end{minipage}
\begin{minipage}[h]{0.43\linewidth}
\begin{tabular}{lccccc}
\toprule
{FGAI}    & {Conv$5_1$}   & {Conv$5_2$}   & {Conv$5_3$}  & {FC6} & {FC7}  \\
\midrule
K-H-GL     & {89.04} 	          & {86.83}                 & \textbf{89.01}       & {70.26}              & {{33.41}}  \\
K-H-LD     & {89.01}	              & {84.08}	                 & {88.09}           & {66.08}	          & {29.22} \\
K-H-SI     & {87.33}	              & {80.72}	                 & {87.08}              & {61.4}	          & {17.54} \\
K-GL-LD    & {\bc{89.15}} &		 {85.9}                     & {88.86}             & {69.83}               & {40.04} \\
K-GL-SI    & {87.4} 	              & {84.94} 	             & {86.9}              & {68.72}              & {26.9} \\
K-LD-SI    & {85.94} 	              & {82.69} 	             & {85.9}              & {65.83}              & {24.72}  \\
H-GL-LD   & \bc{89.15}             & \bc{86.9} 	         & {88.79}                       & {70.36}       & \bc{41.29} \\
H-GL-SI   & {87.54} 	               & {85.33} 	             & {86.76}              & \bc{70.4}            & \textbf{51.11} \\
H-LD-SI   & \textbf{89.81} 	        & \textbf{87.86} 	             & \bc{88.9}          & \textbf{70.9}               & {39.15} \\
GL-LD-SI & {87.4} 	                   & {81.54} 	             & {85.54}              & {58.58}                & {23.26} \\
\bottomrule
\textbf{Mean}     & 89.15	 &  86.83	& 87.58  & 68.11  &  32.66\\
\bottomrule-
\end{tabular}
\end{minipage}
\end{table}

\NW{The comparison between the single GAI and the FGAI with H-LD-SI combination across the different layers of the AlexNet and the Vgg-vd16 networks is reported in Table~\ref{tab:result2-BU4D}. The findings consolidate the results found with the Bosphorus dataset on the superiority of our fusion scheme, in addition to confirming the discrimination power of the convolution and pooling layers compared to the fully connected ones.}
\begin{table}[t]
\caption{BU-4DFE dataset: Classification rates obtained with a single GAI versus FGAI:H-LD-SI for AlexNet and Vgg-vd16.}
\label{tab:result2-BU4D}
\center
\begin{tabular}{lcccccc}
\toprule
{AlexNet}            & { K} & { H} & { GL} & { LD} & { SI}          &  H-LD-SI \\
\midrule
\textbf{Conv4} 		& 63.4	& 72.1 & 61.9 & 75.9 & 77.33 & \textbf{91.34} \\
Conv5               & 61.1  & 65.3 & {60.21} & 72.25 &   70.19     & {86.80} \\
pool5               & 62.7  & 70.90 &  61.63 & 74.32  & {73.43}         & {90.02} \\
FC6                 & 48.12 & 56.81 & 50.38  &  51.40   & {52.01}        & {72.30} \\
FC7                 & 40.71 & 42.44 & 39.89  &   40.92   & {38.01}        & {48.09} \\
\bottomrule
\\
{VGG-vd16} & \multirow{3}{*}{} \\
\midrule
\textbf{Conv5\_1}  & 63.24 & 68.93 & 64.14 & {67.55} & 70.02       & \textbf{89.81} \\
Conv5\_2           & 62.87 & 66.34 & 61.98 & 66.17 & {68.44}              & {87.86} \\
Conv5\_3           & 63.09 & 67.81 & 64.07 & 67.25 & {69.32}              & {88.9} \\
FC6                & 51.04 & 54.11 & 53.38 & 56.39 & {53.01}              & {70.90} \\
FC7                & 28.03 & 30.15 & 29.73 & 30.33 & {31.01}              &  {39.15} \\
\bottomrule
\end{tabular}
\end{table}

\subsubsection{Facial Expression Classification}
\NW{On BU-4DFE, we experimented our proposed method in a static and dynamic mode in order to compare it with different state-of-the-art solutions. In the first mode, we recognized the expression in each frame, whereas in the second mode the decision on the expression is made upon examining a sequence of frames. For the sake of the comparison, we report the experimentation protocol in comparison with representative state-of-the-art methods. In this regard, we note that most of the methods used $60$ subjects in a dynamic mode~\cite{yin2008,SANDBACH2012-IVC,Berretti2013,Amor2014,Xue2015,Zhen2016-ICPR,Yao2018,Zhen2018}, and few considered all the subjects~\cite{Fang2011-ICCVW,real2013,meguid2018,Dapogny2018}. 
For the first group, the experimental protocols show some differences; for example, most of the methods adopted a $10$-fold cross-validation apart of~\cite{SANDBACH2012-IVC}, where a $6$-fold cross-validation was used. Some methods considered the whole video sequence~\cite{Berretti2013,Amor2014,Xue2015,Zhen2016-ICPR,Zhen2018}, while other methods used frames sampled with a sliding-window~\cite{yin2008,real2013}.}
\NW{Methods in the second group~\cite{Fang2011-ICCVW,real2013}, and~\cite{meguid2018,Dapogny2018} adopted a dynamic mode and a static mode, respectively. In~\cite{Fang2011-ICCVW}, a $10$-fold cross-validation was used on the full sequence, whereas in~\cite{real2013} a $15$-frames sequence starting from the first onset frame was considered, and the cross-validation scheme was not reported. In~\cite{Dapogny2018}, a $5$-fold cross-validation protocol was considered with seven classes rather than six (neutral plus six other expressions). The samples in the classes were obtained by selecting $8,219$ frames from all the video, but the way in which the frames were selected was not reported.}

\NW{In our static mode experimentation, we considered seven classes, \emph{i.e.}, the neutral expression plus the six other expressions. Note  that even though the neutral frames are supposed to be at the beginning and the end of the video sequence, according to the order neutral-onset-apex-outset-neutral, we found that this temporal pattern is breached in several samples, therefore we selected these neutral frames manually. For the other expressions, we selected 10 frames taken at the $80^{th}$ estimated as the apex frame. We performed a quality control check to ensure the validity of the frame. Classification has been obtained using $5$-fold cross validation, which makes our setting quite comparable to~\cite{Dapogny2018} and~\cite{meguid2018}. 
Table~\ref{tab:stat_expression-comparison} reports results obtained with our method together with competitive state-of-the-art methods, namely the method of Meguid et al.~\cite{meguid2018} and two methods used by Dapogny et al.~\cite{Dapogny2018}. The former method (method-1) is  the baseline method reported that  employs geometric and HOG descriptors with a standard random forest classifier. The second  method (method-2) employs geometric descriptors and CNN features with the neural forest classifier}. We notice that our proposed method outperforms the state-of-the-art solution by a large margin of 17\% and 15\% for the AlexNet and the Vgg-vd16 versions, respectively.
\begin{table}[htb]
\caption{BU-4FE dataset: Classification rate obtained with the FGAI:H-LD-SI compared with the state-of-the-art methods.}
\label{tab:stat_expression-comparison}
\center
\begin{tabular}{lc}
\toprule
{Method} & {Accuracy} \\
\hline
Abd El Meguid et al.~\cite{meguid2018}   & {73.10} \\
Dapogny et al.~\cite{Dapogny2018}-1       & {72.80} \\
Dapogny et al.~\cite{Dapogny2018}-2       & {74.00} \\
compressed-AlexNet    	                                    & \textbf{{91.34}} \\
compressed-Vgg-vd16                                    & \textbf{{89.81}} \\
\bottomrule
\end{tabular}
\end{table}

\NW{For the dynamic mode experimentation, we considered $60$ subjects, $10$-fold cross validation, and a decision for the expression of a given frame $i$  made using a sliding temporal window of size $6$. The classification is performed by computing the histogram of the expressions in the moving window and selecting the most plausible expression by majority voting. This setting makes our protocol close to its counterparts in~\cite{Berretti2013,Amor2014,Xue2015,Zhen2016-ICPR}. However, a major difference remains. Unlike these methods, our approach does not employ proper temporal features (derived from a frame sequence), despite the fact that the majority voting acts on chronologically ordered frames. Thus our method  does not fully profit from the dynamic aspect of the data. This limitation put our method in a disadvantageous position with respect to the other dynamic methods. 
Table~\ref{tab:bu-4dfe} reports the previous methods, their experimentation protocols and their results compared to those obtained in this paper. Our method scores second, with a classification rate of $95\%$ slightly below the method of Yao et al.~\cite{Yao2018} that uses the key-frames (95.13\%). Compared with methods adopting close protocols, our approach achieved the best performance.}
\begin{table}[ht]
\caption{BU-4DFE dataset: Comparison with state-of-the-art solutions in dynamic mode.  The ``Protocol'' column reports the experimental setting with the following symbols:NUmber of expressions(\#E), number of subjects(\#S), number of folds(\#-CV), frames used (Full-seq., Key-frames, Win=\#) } 
\label{tab:bu-4dfe}
\begin{tabular}{l|l|c}
\multicolumn{1}{l|}{\textbf{Method}} & \multicolumn{1}{l|}{\textbf{Protocol}} & \multicolumn{1}{l}{\textbf{Acc.} (\%)} \\ \hline  \bottomrule
Sun et al.~\cite{yin2008}, 2008 & --, 6E, 60S, 10-CV, Win=6 & 90.44 \\ \hline
Sandbach et al.~\cite{SANDBACH2012-IVC}, 2012 & D, 6E, 60S, 6-CV, Win=4 & 64.60 \\ \hline
Fang et al.~\cite{Fang2011-ICCVW}, 2012 & D, --, 100S, 10-CV, -- & 74.63 \\ \hline
Reale et al.~\cite{real2013}, 2013 & D, 6E, 100S, -, Win=15 & 76.90 \\ \bottomrule
Berretti et al.~\cite{Berretti2013}, 2013 & D, --, 60S, 10-CV, Full-seq. & 79.40 \\ \hline
Ben Amor et al.~\cite{Amor2014}, 2014 & D, --, 60S, 10-CV, Full-seq. & 93.21 \\ \hline
Xue et al.~\cite{Xue2015}, 2015 & D, --, 60S, 10-CV, Full-seq. & 78.80 \\ \hline
Zhen et al.~\cite{Zhen2016-ICPR}, 2016 & D, --, 60S, 10-CV, Full-seq. & 94.18  \\ \hline
Ours                                & D, --, 60S, 10-CV, Full-seq. & \bc{95.00} \\
\bottomrule
Yao et al.~\cite{Yao2018}, 2018 & D, --, 60S, 10-CV, Key-frames & 87.61 \\ \hline
Zhen et al.~\cite{Zhen2018}, 2017 & D, --, 60S, 10-CV, Key-frames & \bf{95.13} \\
\bottomrule
\end{tabular}
\end{table}

\section{Discussion and Conclusions}\label{sect:conclusions}
\NW{In this paper, we presented a novel paradigm for learning effective feature representations from 2D texture and 3D geometric data. We proposed an original scheme for mapping local geometric descriptors, computed on the 3D mesh surface, onto their corresponding textured images. The resulting geometrically augmented images (GAIs) are then combinatorially selected to generate multiple three channel images, forming fused geometrically augmented images (FGAIs). This newly proposed representation  is employed with the standard AlexNet and Vgg-vd16 CNNs in a transfer learning mode to extract discriminative features. In addition, our mesh data down-sampling scheme achieves significant computational efficiency, with a gain of the order of ten, without excessively compromising the performance of the extracted features.}

\NW{The extensive experimentation conducted on AUs and facial expression recognition on two public datasets  evidenced the high discriminative capacity of the learned features derived from the FGAIs and the competitive performance of their counterparts derived from down-sampled data. It also reports on the performance behavior of the extracted features across the different layers of the CNN models. 
We also conducted a comprehensive comparison with state-of-the-art methods of facial expression in static and dynamic settings. 
In the static mode, our method achieved a significant boost in performance exceeding $18\%$ and $17\%$ on the Bosphorus and the BU-4DE datasets, respectively.
In the dynamic mode, and despite being devoid of temporal features, our method reached a slightly lower performance than the best state-of-the-art method. Yet, it showed competitive and better performance compared with methods adopting close experimentation protocols.}

\NW{We believe that several aspects in our method contribute to the achieved significant boost of performance. First, the fusion of 2D and 3D information which allows a face discriminative capacity largely above 
its  individual 2D and 3D counterparts ~\cite{soltana2010}. 
Second, the early-level fusion aspect in our method, whereby 3D geometry and 2D photometric descriptors are fused at pixel level. This is in-line with previous findings confirming that among the four levels of fusions, namely, \textit{data}, \textit{feature}, \textit{score}, and \textit{decision}, the low-level fusion (data and feature) performs better than its higher level counterparts (score and decision)~\cite{ross2003}.
Third, by proposing the so aforementioned early fusion in a CNN network, we implicitly perform a fusion of learned face representation emanating from texture and shape information. This can be viewed as a continuation of the fusion paradigm proposed by Jung et al.~\cite{jung2015} in their 2D facial expression recognition work, where they reported that using a combination of geometric and deep learned texture representations significantly improves the performance. Our method push further this paradigm by fusing both learned shape and texture information.} 

\NW{As future work, there are several directions that can be explored. At first, comes investigating the effect of changing the permutations of the three GAIs on the performance, and check if a specific FGAI permutation performs better than others.  It would be also attempting to go beyond the three GAIs fusion constraint which is imposed by the architecture of the input layer in the AlexNet and Vgg-vd16. For example, one can fusing four or five GAIs at the input level. This, however, requires designing and training a new architecture, which would be time and resource demanding. Besides, this does not align with our advocacy of utilizing pre-trained architecture and profiting from their potentials.} 

\NW{The discrimination capacity results of the features obtained the different layers suggest using the criterion of Eq.~\eqref{equ:descrit}, as a feature selection tool, allowing at the same time dimensionality reduction. Even though the features taken at the layers are not necessarily expected to have a Gaussian distribution, bearing in mind the non-linear activation functions across the networks and the output sparsity in the layers~\cite{lorieul2016}, we believe it would be worth investigating adopting  this criterion as a dimensionality reduction tool. 
Finally, we plan to enlarge the scope of our framework and extend it to other categories of mesh surfaces for medical and remote sensing applications.}

\bibliographystyle{IEEEtran}
\end{document}